\documentclass[11pt]{article}

\usepackage{acl}

\usepackage{times}
\usepackage{latexsym}

\usepackage[T1]{fontenc}
\usepackage[utf8]{inputenc}

\usepackage{microtype}

\usepackage{inconsolata}

\usepackage{graphicx}

\usepackage{hyperref}
\usepackage{url}
\usepackage{booktabs} 
\usepackage{tabularx}
\usepackage{multirow}
\usepackage{subcaption}
\usepackage{caption}
\usepackage{amssymb,amsmath}
\usepackage{xspace}
\usepackage{pifont}
\usepackage{array}

\usepackage[most]{tcolorbox}
\usepackage{xcolor}

\usepackage{makecell} % 可选：更好看的列标题换行
\newcolumntype{C}[1]{>{\centering\arraybackslash}p{#1}}

\newcommand{\cmark}{\ding{51}}  % ✓
\newcommand{\xmark}{\ding{55}}  % ✗

\usepackage{enumitem}
\usepackage[table]{xcolor} % 必须带 [table] 参数才能给单元格上色

\definecolor{lowGreen}{HTML}{E8F5E9}   % 极浅绿 (用于低频转换)
\definecolor{midGreen}{HTML}{A5D6A7}   % 中柔绿
\definecolor{highGreen}{HTML}{66BB6A}  % 莫兰迪绿 (高频/核心贡献)
\definecolor{lowRed}{HTML}{FFEBEE}     % 极浅红 (用于微小误伤 2.8%)
\definecolor{midRed}{HTML}{EF9A9A}     % 柔和红
\definecolor{highRed}{HTML}{E57373}    % 哑光红 (用于持续错误)

\title{A Training-Free Regeneration Paradigm: Contrastive Reflection Memory Guided Self-Verification and Self-Improvement}
\author{
  \textbf{Yuran Li}$^{1}$, \textbf{Di Wu}$^{1,2}$, \textbf{Benoit Boulet}$^{1}$ \\
  $^1$Department of Electrical and Computer Engineering, McGill University \\
  $^2$Department of Mechanical, Industrial and Aerospace Engineering, Concordia University\\
  \texttt{yuran.li@mail.mcgill.ca}; \texttt{di.wu5@mcgill.ca}; \texttt{benoit.boulet@mcgill.ca}
}

\begin{document}
\maketitle
\begin{abstract}
Verification-guided self-improvement has recently emerged as a promising approach to improving the accuracy of large language model (LLM) outputs. 
However, existing approaches face a trade-off between inference efficiency and accuracy: iterative verification–rectification is computationally expensive and prone to being trapped in faulty reasoning, while best-of-N selection requires extensive sampling without addressing internal model flaws.
We propose a training-free regeneration paradigm that leverages an offline-curated contrastive Reflection Memory (RM) to provide corrective guidance, while regenerating from scratch helps break out of faulty reasoning. At inference time, the method performs RM-guided self-verification followed by a single RM-guided regeneration, avoiding both iterative correction and multi-sample selection.
We evaluated our method on nine benchmarks that span algorithmic, reasoning, symbolic, and domain-specific tasks in both small- and large-scale LLMs. Experiment results show that our method outperforms prior methods while maintaining low computational cost. Code is available at \url{https://anonymous.4open.science/r/Supplementary-code-5B17}.

\end{abstract}

\section{Introduction}
Scaling language models to billions of parameters has enabled them to encode broad domain knowledge and exhibit strong reasoning abilities. Nevertheless, ensuring the correctness of generation remains challenging. Beyond improving reasoning generation alone, recent work explores verification-guided improvement. However, without ground-truth supervision, large language models (LLMs) still struggle to reliably assess their reasoning. This gap between generation, verification, and improvement remains a major obstacle to further advancing LLMs' performance~\cite{kamoi2024can, huang2024cannot, tie2025correctbench}.

Extensive research has investigated the improvement in LLMs reasoning generation. \citet{wei2022chain} proposed chain-of-thought (CoT) prompting. Building on this, \citet{wang2023self} introduced the self-consistency decoding strategy.
Subsequently, more advanced frameworks such as Tree-of-Thought (ToT), Algorithm-of-Thought (AoT), and Forest-of-Thought (FoT) \cite{yao2023tree, sel2024algorithm, bi2025forest} were developed, which improve reasoning capabilities by encouraging explicit search over intermediate steps. 
Additionally, numerous studies have shown that the generation quality can be further improved with well-designed few-shot demonstrations. \citet{zhang2023automatic} introduced automatic demonstration generation and \citet{chia2023contrastive, mo2024cicl, gao2024customizing} proposed contrastive-example In-Context-Learning (ICL).
However, these methods primarily focus on reasoning generation, leaving verification and rectification underexplored.

% \begin{figure}
%     \centering
%     \includegraphics[width=0.8\linewidth]{./pic/introduction.pdf} 
%     \caption{Comparison of three verification-guided improvement paradigms.}
%     \label{fig:introduction} 
% \end{figure}

\begin{table*}[t]
\scriptsize % 比 \small 更紧凑
\centering
\setlength{\tabcolsep}{3pt} % 缩小列间距
\renewcommand{\arraystretch}{1.0} % 行距微调
\begin{tabular}{p{3.8cm} >{\centering\arraybackslash}p{1.8cm} >{\centering\arraybackslash}p{2.5cm} >{\centering\arraybackslash}p{3cm} >{\centering\arraybackslash}p{1.5cm}}
\toprule
\textbf{Method} & \textbf{Learning} & \textbf{Verification} & \textbf{Improvement Method} & \textbf{Iterate} \\
\midrule
\multicolumn{5}{l}{\emph{Prompt Engineering}} \\
\midrule
Self-Refine~\cite{madaan2023self} & ICL & self-critic & Explicit FB Rectify  & \cmark \\
Reflexion~\cite{shinn2023reflexion} & ICL & Oracle & Explicit FB Rectify  & \cmark \\
ProCo~\cite{wu2024large} & ICL & Solve Verif. Q. & Implicit Rectify & \cmark \\
ST CoT~\cite{wu2025rethinking} & ICL & Uncertainty-guided & Implicit Rectify  & \cmark \\
\textbf{Ours} & ICL & RM-guided & Regenerate &  \xmark \\
\midrule
\multicolumn{5}{l}{\emph{External Tool}} \\
\midrule
CRITIC~\cite{gou2024critic} & ICL & Tool-Assist & Explicit FB Rectify & \cmark \\
Self-Debugging~\cite{chen2024selfdebugging} & ICL & Programmatic & Explicit FB Rectify &  \cmark \\
\midrule
\multicolumn{5}{l}{\emph{Training-enabled}} \\
\midrule
V-STaR~\cite{hosseini2024vstar} & Fine-tuning & Trained Verifier & Best-of-N &  \cmark \\
ReflectEvo~\cite{li2025reflectevo} & Fine-tuning & Oracle & Implicit Rectify &  \xmark \\
SCoRe (RL)~\cite{kumar2025score} & RL fine-tuning &\xmark & Implicit Rectify &  \xmark \\
\bottomrule
\end{tabular}
\caption{Comparison across major verification-guided improvement methods. 
\cmark/\xmark~indicate presence/absence of a component; “Oracle” denotes access to ground-truth labels; “Rectify” vs. “Regenerate” indicate whether prior outputs are reused; “Explicit” vs. “Implicit” indicate whether textual feedback is produced.}
\label{tab:self-correction-comparison}
\end{table*}

A complementary line of research targets reflection and refinement to further improve LLM performance, as summarized in Table~\ref{tab:self-correction-comparison}.
Training-free approaches such as Self-Refine \cite{madaan2023self} rely on LLM to self-verify, while Reflexion \cite{shinn2023reflexion} uses ground-truth labels for verification. These methods iteratively refine output using LLM-generated feedback. CRITIC~\cite{gou2024critic} and Self-Debugging~\cite{chen2024selfdebugging} focus on specific tasks (like algorithmic reasoning, code debugging, or knowledge-based QA), using external tools such as a search engine, a calculator, or programmatic execution to verify, then rectifying iteratively with LLM-generated feedback. Self-Train CoT (ST CoT)~ \cite{wu2025rethinking} instead uses entropy-based uncertainty for verification and implicitly refines responses by prompting the LLM to “think deeper” through adaptive iterations.
In addition, fine-tuning–based methods have also been explored. V-STaR~\cite{hosseini2024vstar} trains a verifier to select the best response from multiple candidates. ReflectEvo~\cite{li2025reflectevo} relies on supervised verification to trigger reflection and rectification. SCoRe \cite{kumar2025score} removes explicit verification but enforces a second refinement pass, without guaranteeing error correction.

% These approaches exhibit several limitations as shown in Figure~\ref{fig:introduction}. 
Despite recent progress, existing approaches still exhibit several limitations in achieving practical, reliable, and efficient verification-guided improvement.
First, many self-refinement methods, such as ReflectEvo \cite{li2025reflectevo}, Reflexion \cite{shinn2023reflexion}, CRITIC~\cite{gou2024critic}, and Self-Debugging~\cite{chen2024selfdebugging} rely on ground-truth or external tools for verification, which are impractical for general inference.
Second, best-of-$N$ selection methods, like V-STaR~\cite{hosseini2024vstar} leave internal flaws unaddressed when all responses have errors. Although iterative approaches could mitigate this issue but introduce substantial computational overhead. 
Third, iterative refinement approaches, such as Self-Refine \cite{madaan2023self}, ST CoT \cite{wu2025rethinking} perform unsupervised multi-step refinement, accumulating errors in feedback and rectification while incurring high computational cost.
These limitations motivate a rethinking of verification-guided improvement: how to enable reliable self-verification and refinement without ground-truth supervision, while avoiding costly iterative generation thereby reducing inference cost and improve real-world applicability.

To address these challenges, we introduce Contrastive Reflection Memory (RM), which distills the corrective insights of a stronger teacher model into a structured repository of base-model errors (Sec.\ref{subsec:RM construction}). 
During inference, relevant RM entries are retrieved (Sec.\ref{subsec: retrieval}) and used in two training-free variants: (i) RM-primed prompting, which uses RM to jointly perform verification and rectification in one forward pass (Sec.\ref{subsec:Direct Performance Boost}); and (ii) RM-guided self-verification and regeneration (Sec.\ref{subsec:Verification and Regeneration}).
Extensive experiments across multiple LLMs demonstrate the applicability of our method. Without fine-tuning, external tools, or ground-truth label, our method achieves the best performance with lower computational cost (Sec. \ref{sec:Experiment}).
In summary, our main contributions are:

\begin{itemize}[leftmargin=0pt]
\setlength{\itemsep}{1pt}      % 每个 \item 之间的垂直间距
\setlength{\parskip}{1pt}      % 段落之间
\item \textbf{Contrastive Reflection Memory Curation.} We introduce offline-curated RM that distills corrective reasoning insights from a teacher model into principles and contrastive examples. This enables base models to leverage teacher-level heuristics for inference-time self-improvement.
\item \textbf{Enhanced Self-verification.} We introduce a verifier guided by retrieved RM and model uncertainty to provide reliable verification signals without requiring ground-truth labels or external tools.
\item \textbf{Efficient Single-step Regeneration.} We propose an RM-guided regeneration paradigm that replaces iterative editing with a single regeneration step, improving reasoning reliability while achieving a better cost–performance trade-off than sampling-heavy or multi-refinement methods.

\end{itemize}

\section{Training-Free Regeneration Paradigm}
\label{sec:method}

\begin{figure*}[htbp]
    \centering
    \includegraphics[width=0.85\linewidth]{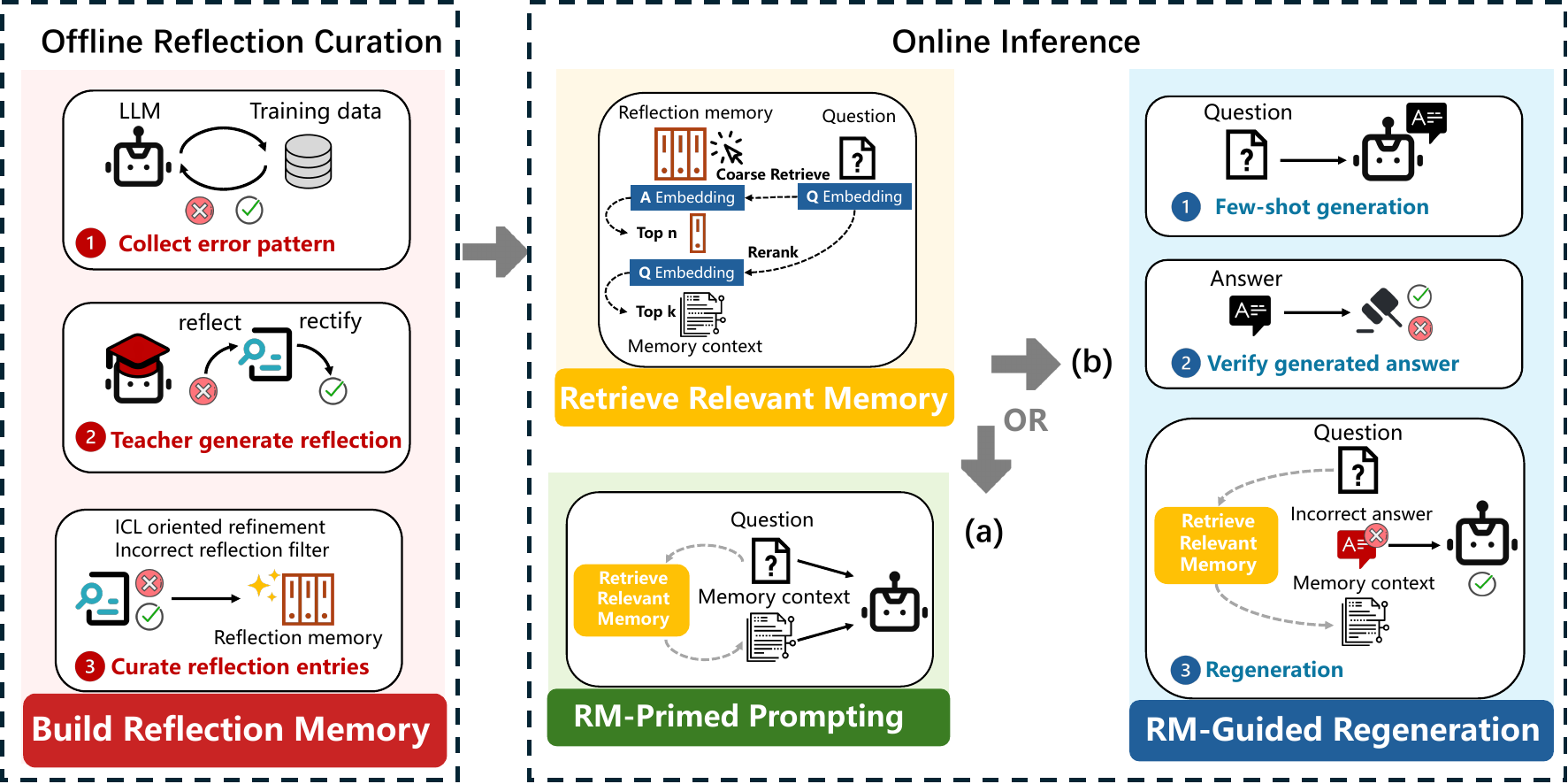} 
    \caption{Overview of our training-free regeneration framework.
Left: Offline Reflection Curation.
Middle: Online Retrieval.
Right: Online Inference. (a) Direct Performance Boost: the LLM answers the query conditioned on the retrieved memory context. (b) RM-Guided Regeneration: the LLM first generates and verifies an answer; if incorrect, we retrieve reflections and regenerate a corrected answer conditioned on the memory context.}
    \label{fig:framework} 
\end{figure*}

As illustrated in Figure~\ref{fig:framework}, our method operates in two stages.
\textbf{(1) Offline Reflection Memory Curation.} First, we collect erroneous reasoning patterns from training data. A teacher LLM then reflects on and rectifies these errors. The resulting reflections and rectifications are subsequently refined and filtered to construct a structured contrastive RM.
\textbf{(2) Online Inference.} At inference time, relevant RM are retrieved based on embedding similarity and can be used in two variants: (i) RM-Primed, RM-guided pre-reasoning primed prompting or (ii) RM-Regen, RM-guided post-reasoning verification and regeneration.

\subsection{Offline Reflection Memory Curation}
\label{subsec:RM construction}
This section details the curation of contrastive RM from base-model errors, teacher reflections, and an ICL-oriented refinement and filtering pipeline.

\subsubsection{Error Pattern Collection}
Let $Q$ denote the questions in the training set. For each $q \in Q$, we run the base LLM for $k$ times and obtain responses: $M_{base}(q) = \{\, a_1, a_2, \dots, a_k \}$.
When all responses match the ground truth, we label the pair $(q,a_1)$ as a correct demonstration (we take the first instance without loss of generality).
If responses are inconsistent, it indicates the model has the required knowledge but exhibits stochastic reasoning variance. In such cases, we retain the response $a_i$ that matches the ground truth and label $(q,a_i)$ as a clean supervision signal, filtering out incorrect attempts to provide an unambiguous reference that reduces uncertainty and reinforces reliable reasoning paths.
When all responses are incorrect, we label $(q,a_1)$ as an error-prone case, reflecting potential deficiencies in the model’s reasoning or knowledge.

% It is written as equation~(\ref{eq:error_collection}):

% {\small
% \begin{equation}
% \label{eq:error_collection}
% (a, v) =
% \begin{cases}
% (a_1, \text{correct}), 
% & \text{if } \forall i,\ \mathrm{Ans}(a_i) = y^{\text{gt}}, \\[4pt]
% (a_j, \text{correct}), 
% & \text{if } \exists j,\ \mathrm{Ans}(a_j) = y^{\text{gt}}, \\[4pt]
% (a_1, \text{incorrect}),  
% & \text{if } \forall i,\ \mathrm{Ans}(a_i) \neq y^{\text{gt}}.
% \end{cases}
% \end{equation}
% }
% \noindent
% where $a$ include reasoning and final answer, $\mathrm{Ans}(a)$ denote the final answer extracted from $a$.

\subsubsection{Teacher Reflection Generation}
A higher-capability LLM is employed as a teacher model to analyze erroneous reasoning $a^{-}$, reflect underlying mistakes $r$, and provide corrected reasoning traces $a^{+}$:  $(r, a^{+}) = M_{\mathrm{teacher}}\big(q, a^{-}; \tau_{\mathrm{reflect}}\big)$.
In addition, the teacher model is prompted to derive concise and generalizable principles $\pi$ from the reflections, which can help improve future reasoning performance: $\pi = M_{\mathrm{teacher}}\big(r; \tau_{\mathrm{principle}}\big)$.
The prompt template $\tau_{\mathrm{principle}}$, $\tau_{\mathrm{reflect}}$ used for teacher-guided reflection are provided in Appendix~\ref{sec:reflection prompt} and Appendix~\ref{sec:principle prompt}.

\subsubsection{Filtering and Refinement}
Directly using the collected pairs of incorrect reasoning, reflection, and rectification is suboptimal.
Incorrect reasoning is often verbose, inconsistent with the rectified reasoning format, and may contain compounded errors, which reduces the effectiveness of ICL.
Moreover, teacher-generated reflections and rectifications can be unreliable.
To address these issues, we introduce a filtering step to remove cases with incorrect rectifications, and an ICL-oriented refinement step to clean, normalize, and structure these examples for in-context use.

\textbf{ICL-Oriented Refine}~ Motivated by rewrite-then-regenerate \cite{shao2023synthetic} and reverse-engineered reasoning \cite{wang2025reverse}, we propose ICL-oriented refinement, where the teacher LLM rewrites each incorrect response $a^-$ into a compact contrastive negative $\hat a^-$ by removing verbosity, preserving the incorrect final answer, and explicitly surfacing the specific error identified in reflection: $\hat a^- = M_{\mathrm{teacher}}(a^-, r, a^+; \tau_{\mathrm{refine}})$. The refined prompt template $\tau_{\mathrm{refine}}$ is provided in Appendix~\ref{sec:refine prompt}.
This refinement concentrates the error signal and sharpens the contrast between incorrect and corrected reasoning, improving contrastive ICL effectiveness.
To conceptually illustrate the intuition, we define the "information contribution" of a refined negative and its reflection as:

{\small
\setlength{\abovedisplayskip}{3pt}
\setlength{\belowdisplayskip}{3pt}
\setlength{\abovedisplayshortskip}{3pt}
\setlength{\belowdisplayshortskip}{3pt}
\begin{equation}
\label{eq:ICL-information}
\mathcal{I}(\hat a^-; r, a^+)
\approx
p_{M_{\mathrm{base}}}
\big(
    y^{*} \mid 
    q',\,
    \mathrm{ICL}(q, \hat a^-, r, a^+)
\big).
\end{equation}
}
where $\mathcal{I}(\hat a^-; r, a^+)$ denotes the base model’s prob-
ability of predicting the correct answer $y^{*}$ for a test
question $q'$ when prompted with the contrastive
demonstration.

\subsubsection{Reflection Memory Structuring}

Finally, we organize the curated contrastive RM into two subsets, correct entries $\mathcal{R}^{+}$ (validated solutions) and incorrect entries $\mathcal{R}^{-}$ (reflections and rectifications), as follows:

{\small
\setlength{\abovedisplayskip}{3pt}
\setlength{\belowdisplayskip}{3pt}
\setlength{\abovedisplayshortskip}{3pt}
\setlength{\belowdisplayshortskip}{3pt}
\begin{align}
\mathcal{RM} &= \mathcal{R}^{+} \cup \mathcal{R}^{-}, \\
\mathcal{R}^{+} &= \{\, (q,a^+,v)\;|\; v=\text{correct}\,\}, \\
\mathcal{R}^{-} &= \{\, (q,\hat{a}^-,v,r,a^{+},\pi)\;|\; v=\text{incorrect}\,\}.
\end{align}
}
Here $q$ denotes question; $\hat{a}^-$ is refined incorrect answer;
$v$ is verification;
$r$ is reflection;
$a^{+}$ is corrected answer; and
$\pi$ is the principle summarized from reflection.

% We define 10 coarse-grained error categories covering both reasoning and factual dimensions:
% {Calc, Logic, Boundary, Assumption, Format, Context, Grounding, Redundancy, Knowledge, Other}.
% This taxonomy balances coverage and tractability, ensuring sufficient diversity while maintaining consistent annotation quality.
% We found that finer-grained taxonomies yielded marginal improvement but increased labeling inconsistency.

% \begin{figure*}[htbp]
%     \centering
%     \includegraphics[width=0.7\linewidth]{./pic/gpt3.5_gsm_rewrite_bar.pdf} 
%     \caption{xxxxx}
%     \label{fig:xxxxx} 
% \end{figure*}

\subsection{Online Inference}
In this section, we describe how RM is retrieved during inference and how it is used in two variants.

\subsubsection{Reflection Memory Retrieval} %memory activation
\label{subsec: retrieval}
At inference time, we use a retrieval-then-reranking pipeline to select the most relevant RM. First, we use Contriever \cite{izacard2021unsupervised}, a pretrained contrastive-learned dense retriever, to retrieve a coarse top-$n$ candidate set by embedding similarity between the test query $q'$ and $a^+$ in RM \cite{you2024llm}.  
We then use MPNet-based sentence encoder ~\cite{song2020mpnet} to rerank these candidates by semantic similarity between $q'$ and $q$ in RM, and retain the top-$k$ entries as the final memory context.
This combines Contriever for broad, reasoning-oriented retrieval with MPNet for fine-grained question matching.

Following \citet{huang2024context}, our RM retrieval-augmented inference can be interpreted through a reinforcement learning (RL) perspective without explicitly performing RL training: a model’s output distribution under a given context resembles a context-dependent policy. Specifically, the task context $q'$ plays the role of state $S$, and the generated answer $a'$ corresponds to action $A$. 
By retrieving the relevant RM demonstrations, the model conditions on previous validation solutions and rectification strategies via ICL, thereby guiding the conditional distribution toward more effective outputs.
Under this interpretation, the implicit policy induced by memory-guided inference can be approximated as:

{\small
\setlength{\abovedisplayskip}{3pt}
\setlength{\belowdisplayskip}{3pt}
\setlength{\abovedisplayshortskip}{3pt}
\setlength{\belowdisplayshortskip}{3pt}
\begin{equation}
\label{eq:policy}
\pi(A=a' \mid S=q')
\;\approx\;
P\big(a' \mid q',\, \mathcal{R}_{\mathrm{top}}^{+} \cup  \mathcal{R}_{\mathrm{top}}^{-}\big),
\end{equation}
}
where $\mathcal{R}_{\mathrm{top}}^{+}$ and $ \mathcal{R}_{\mathrm{top}}^{-}$ denotes retrieved top-$k$ memory contexts from $\mathcal{R}^{+}$ and $\mathcal{R}^{-}$.

\subsubsection{RM-Primed Prompting}
\label{subsec:Direct Performance Boost}

In RM-Primed Prompting, the LLM implicitly performs reasoning, verification, and rectification in a single forward pass, guided by retrieved RM context.
Without memory guidance, the model directly generates an answer based solely on the query \(q'\).
With RM guidance, generation is additionally conditioned on retrieved contrastive memories:

{\small
\setlength{\abovedisplayskip}{3pt}
\setlength{\belowdisplayskip}{3pt}
\setlength{\abovedisplayshortskip}{3pt}
\setlength{\belowdisplayshortskip}{3pt}
\begin{equation}
\label{eq:direct_boost}
P\big(a' = y^* \mid q',\, \mathcal{R}^{+}_{\mathrm{top}} \cup \mathcal{R}^{-}_{\mathrm{top}}\big),
\end{equation}
}
where $y^*$ denotes ground truth answer. 
This indicates that RM shifts the conditional distribution of the model, 
providing actionable signals that boost performance without additional fine-tuning.

% (1) ICL
% Essence of In-Context Learning (ICL)
% ICL is essentially conditional probability modeling: the model leverages the attention mechanism to “match” the input with the provided demonstrations in the context, treating the demonstration patterns as part of the conditional distribution. At its core, it is closer to prompt-driven retrieval and analogy.

%(2) fine-tune
%Essence of Fine-tuning
%Fine-tuning is essentially gradient-based parameter learning: through backpropagation, it updates the model parameters so that the distribution  $P_\theta(y|x)$ better aligns with the target task distribution. At its core, it represents retraining with long-term memory consolidation.

\subsubsection{RM-Guided Verification and Regeneration}
\label{subsec:Verification and Regeneration}
Since some tasks can be solved correctly without additional context, injecting RM may be redundant and can even disrupt an initially correct reasoning process.
In the RM-Guided Regen variant, we explicitly separate verification and regeneration, then trigger regeneration only for responses verified as incorrect.

\textbf{RM–Guided Verification}~ Unlike methods that rely on ground truth or a stronger external verifier, we propose an RM-guided self-verification scheme. Guided by retrieved RM, the same base model first compares its answer with relevant RM entries to detect errors in high precision: $v_{init}=M_{\mathrm{base}}(q',a', \mathcal{R}_{\mathrm{top}}^{+} \cup \mathcal{R}_{\mathrm{top}}^{-}; \tau_{\mathrm{verify}})$. where $v_{\mathrm{init}} \in \{\text{correct}, \text{incorrect}\}$. Then an entropy-based uncertainty check is applied to recover recall by flagging doubtful “correct” answers: $v
= \mathbb{I}\!\left[    H\big(q', a') > \gamma  \right]$. This combination of memory-based filtering and uncertainty-based recovery yields a favourable precision–recall trade-off without fine-tuning, ground truth, or an external judge.

\textbf{RM–Guided Regeneration}~ Responses verified as incorrect are then regenerated from scratch, conditioned on the retrieved RM as:  $a''=M_{\mathrm{base}}( q', \mathcal{R}^{+}_{\mathrm{top}} \cup \mathcal{R}^{-}_{\mathrm{top}}\big)$. 
Unlike RM-Primed, regeneration is invoked only after explicitly verified as incorrect, preventing unnecessary interference with correct reasoning while focusing corrective computation on genuinely erroneous cases.

\section{Experiments}
\label{sec:Experiment}
\subsection{Experiment Setup}
We evaluate our method on diverse benchmarks and models, focusing on overall accuracy, verification quality, and efficiency.

\textbf{Datasets}~ 
(i) \emph{Algorithmic}: GSM8K~\cite{cobbe2021training}, GSM-Hard~\cite{shu2024hardgsm8k}, and MATH (Level 3 subset)~\cite{hendrycks2021math};
(ii) \emph{Common-sense reasoning}: StrategyQA~\cite{geva2021strategyqa} and Bamboogle~\cite{press2023measuring};
(iii) \emph{Symbolic reasoning}: Coin Flip and Letter tasks~\cite{wei2022chain};
and (iv) \emph{Domain-specific tasks}: LegalBench (LB)~\cite{guha2023legalbench} and News Headline (HL)~\cite{Sinha2023headline}.

\textbf{Baselines}~
For RM–Primed, which performs a single forward pass, we compare to Contrastive CoT (paired positive–negative demonstrations). For RM–Regen, which involves generation, verification, and regeneration, we compare with: Reflexion (reflection with episodic memory), ReflectEvo (fine-tuning on reflection-augmented data), Best-of-N (selection among candidates), Self-Refine (iterative self-assessment and refinement),
ProCo (verification-question solving and error-guided rectification), and ST-CoT (uncertainty-guided deeper CoT).

\textbf{Models}~
We conduct experiments with three base models: a large-scale model, GPT-3.5, and two smaller open-weight models, Llama-3.1-8B, Gemma-2-9B, which represent diverse architectural families; a higher-capability teacher model: GPT-5mini.
%To further demonstrate its generalizability, we also test it on additional LLMs, with detailed results reported in the Appendix~\ref{xxx}.
Additional implementation details, including dataset splits, inference and hyperparameter settings, are provided in the Appendix~\ref{sec: implementation detail}.

\subsection{RM-Primed Prompting}

\begin{table}[htbp]
\centering
\scriptsize
\setlength{\tabcolsep}{4pt}  % 列间距稍微缩一点
\renewcommand{\arraystretch}{1} % 行距
\begin{tabularx}{\columnwidth}{ll *{5}{c}}
\toprule
&\textbf{Method} & GSM-H & Bamb & Coin & LB & avg\% \\
\midrule

\multirow{5}{*}[-2pt]{\rotatebox{90}{GPT-3.5}}
&Few-shot CoT           & 41.25 & 55.00 & 74.25 & 44.21 & 61.15 \\
&Contrastive CoT (w/o r) & 39.40 & 49.00 & 59.00 & 35.75 & 57.01 \\
&Contrastive CoT (w r)   & 38.00 & 50.00 & 69.25 & 55.79 & 60.33 \\
&RM-Primed ($\mathcal{R}^{+}$)& 42.50 & 59.00 & 78.00 & 64.21 & 64.99 \\
&RM-Primed                    & \textbf{44.60} & \textbf{62.00} & \textbf{79.75} & \textbf{67.37} & \textbf{66.01} \\
\midrule
\multirow{5}{*}[-2pt]{\rotatebox{90}{Llama3-8B}}
&Few-shot CoT            & 31.80 & 51.00 & 65.75 & 48.42 & 64.19 \\
&Contrastive CoT (w/o r) & 33.20 & 25.00 & 60.75 & 57.89 & 59.56 \\
&Contrastive CoT (w/ r)   & 33.00 & 29.00 & 66.25 & 53.68 & 61.13 \\
&RM-Primed ($\mathcal{R}^{+}$)& \textbf{35.80} & 51.00 & 78.75 & 62.11 & 65.94 \\
&RM-Primed                    & 34.20 & \textbf{54.00} & \textbf{80.75} & \textbf{64.21} & \textbf{68.23} \\
\bottomrule
\end{tabularx}
\caption{Accuracy of different direct-boost methods on representative benchmarks and overall average accuracy. RM-Primed ($\mathcal{R}^{+}$) denotes that only correct entries are used. w/ r and w/o r indicate with and without reflection.}
\label{tab:direct-boost-compact}
\end{table}

We first evaluate RM-Primed against baselines without explicit verification or rectification, as shown in Table~\ref{tab:direct-boost-compact}.
For GPT-3.5, our approach obtains the highest average accuracy of 66.01\%, yielding a 4.86\% point improvement over Few-shot CoT. For LLaMA 3.1-8B, our method reaches an average accuracy of 68.23\%, a 4.04\% gain compared to Few-shot CoT. Moreover, contrastive RM is more effective than RM with only correct entries.
Compared with contrastive CoT, we observe that: (i) simply adding fixed contrastive examples is ineffective; retrieving the most relevant contrastive examples is crucial for consistent gains across benchmarks, and (ii) contrastive examples augmented with reflections are more beneficial than those without. Overall, these results indicate that our RM-primed prompting provides a stronger direct boost than both standard CoT and contrastive CoT baselines. Complete results on nine benchmarks are provided in Appendix~\ref{sec:full results direct boost}, showing the same trend.

% Looking at individual tasks, our methods are competitive or superior on most datasets. For GPT-3.5, our method achieves the best or near-best accuracies on GSM8K, GSM-Hard, StrategyQA, Bamboogle, Coin, LB, and Headline. The largest improvements appear on more challenging or domain-oriented benchmarks such as LB and Bamboogle, where our method substantially outperforms previous direct-boost baselines, suggesting that reflection memory is particularly beneficial when reasoning requires integrating multiple cues or domain knowledge. For LLaMA 3.1-8B, our method dominates on most benchmarks, including GSM8K, StrategyQA, Bamboogle, Coin, Letter, LB, and Headline. While One-shot CoT attains slightly higher accuracy than ours on MATH, this is offset by consistent gains on reasoning and symbolic tasks, leading to the highest overall average.

\begin{table*}[htbp]
\centering
% 更紧凑的表格排版
\scriptsize
\setlength{\tabcolsep}{5pt}  % 列间距
\renewcommand{\arraystretch}{1.2} % 行距
\begin{tabularx}{0.83\textwidth}{cl *{10}{c}}
\toprule
& & \multicolumn{3}{c}{\textbf{Algorithmic}} & \multicolumn{2}{c}{\textbf{Reasoning}} & \multicolumn{2}{c}{\textbf{Symbolic}} & \multicolumn{2}{c}{\textbf{Domain}} & \multicolumn{1}{c}{\textbf{Overall}} \\
\cmidrule(lr){2-12}
\textbf{Model} & \textbf{Method} & GSM8K & GSM-H & MATH & SQA & Bamb & Coin & Letter & LB & HL & avg\%\\
\midrule
\multirow{4}{*}[4pt]{\rotatebox{90}{GPT-3.5}}
% &Reflexion (2 iters)      & 84.00         & 44.00         & 59.00         & 67.75         & 63.00         &{87.25}& 76.00         & 62.11 & 72.00&67.87\\
&Reflexion(3 iters)      & 85.60         & 45.20         & 62.25         & 71.00         & 66.00         &\textbf{88.25}& 78.67         &\textbf{85.26} & 72.00&70.65\\
&Best-of-N (N=3) & 84.60&	46.80&	\textbf{64.00}&	71.50&	63.00&	77.00&	74.67&	32.63&	70.00& 67.29\\
% &RM-Regen ($\mathcal{R}^{+}$) & 88.20         & 64.20         &\textbf{64.60} &\textbf{78.75} & 72.00         & 85.25 & 84.67         & 71.58 & 71.00&75.74\\
&RM-Regen                    & \textbf{90.20}& \textbf{68.60}& 63.40         &\textbf{77.25}         &\textbf{73.00} & 85.75 & \textbf{87.33}&{75.79} &\textbf{73.00}&\textbf{76.94}\\
\midrule

\multirow{4}{*}{\rotatebox{90}{Llama3.1-8B}}
% &Reflexion (2 iters)      & 88.60 & 34.40         & 75.40         & 80.75         & 59.00         & 75.50         & 76.67         &{78.95} & 68.00&70.46\\
&Reflexion(3 iters)      & 89.60 & 36.20         & 77.80         &\textbf{85.00} & 60.00         & 81.25         & 80.67         &\textbf{81.05}         & 70.00&73.26\\
&ReflectEvo                & 85.40 & 32.00         & 73.60         & 71.25         & 53.00         & 67.75         & 68.67         & 50.53         & 62.00&64.74\\
&Best-of-N (N=3) & 89.60&	34.20&	76.80&	77.50&	53.00&	72.00&	70.67&	64.21&	75.00& 69.08\\
% &RM-Regen ($\mathcal{R}^{+}$) & 91.80 & 38.20         & 79.40         & 81.50         & 60.00         & 85.00         & 86.67         & 65.26         & 74.00&74.28\\
&RM-Regen             &\textbf{92.60}&\textbf{41.00} &\textbf{80.60} &{83.25}&\textbf{63.00} & \textbf{85.75}&\textbf{89.33} & {66.32}&\textbf{75.00}&\textbf{75.85}\\

\midrule
\multirow{4}{*}{\rotatebox{90}{Gemma2-9B}}
% &Reflexion (2 iters)  &91.00 &50.20 &67.80 &62.75 &49.00 &59.00	&47.33	&74.74	&71 &\\
&Reflexion(3 iters)   &91.20 &50.80	&72.00 &67.50 &50.00 &60.00	&\textbf{51.33}	&75.79	&71.00 & 67.40\\
&ReflectEvo           &85.80 &44.40 &63.80 &65.00 &47.00 &67.25 &42.14 &48.42 &72.00 &62.92 \\
&Best-of-N (N=3)      &88.80 &48.40 &67.20 &53.75 &46.00 &49.00 &34.67 &45.26 &67.00 &59.78\\
&RM-Regen             &\textbf{92.00} &\textbf{53.00} &\textbf{72.20} &\textbf{73.50} &\textbf{53.00} &\textbf{78.25} &44.29 &\textbf{78.95} &\textbf{73.00} & \textbf{71.42}\\
\bottomrule
\end{tabularx}
\caption{Accuracy under oracle verification (verification-and-rectification/selection baselines vs. ours).}
\label{tab:with oracle verification}
\end{table*}

\begin{figure*}[htbp]
    \centering
    \begin{subfigure}{0.245\linewidth}
        \centering
        \includegraphics[width=\linewidth]{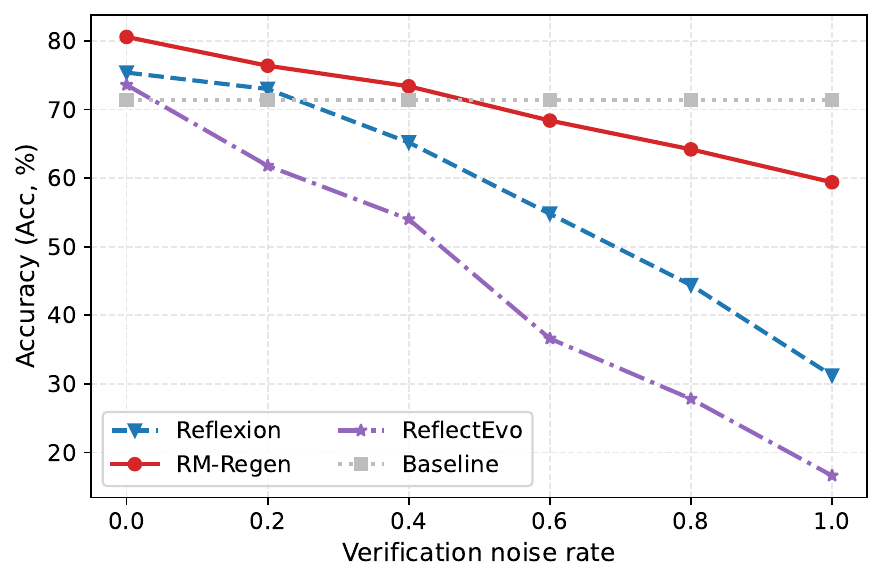}
        \caption{MATH}
        \label{fig:verification_noise_math}
    \end{subfigure}
    \hfill
    \begin{subfigure}{0.245\linewidth}
        \centering
        \includegraphics[width=\linewidth]{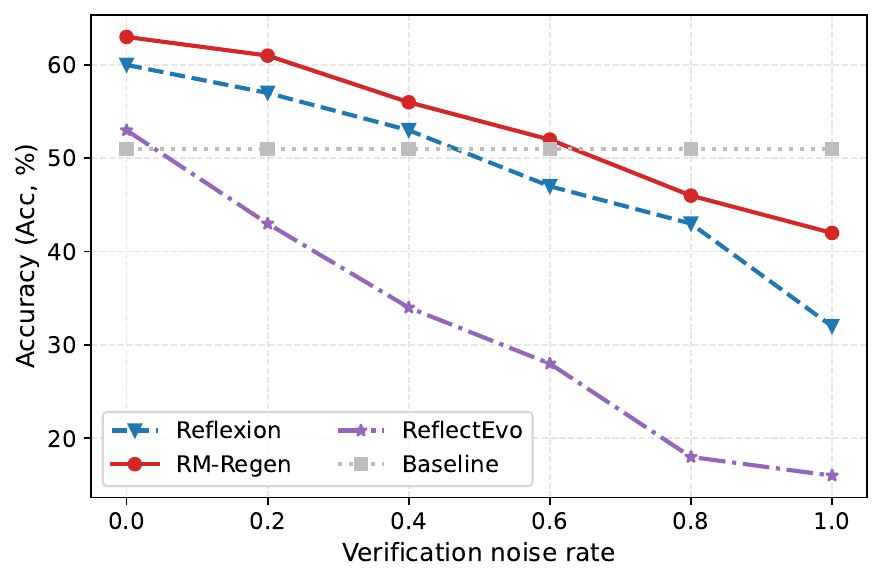}
        \caption{Bamboogle}
        \label{fig:verification_noise_bamboogle}
    \end{subfigure}
    \hfill
    \begin{subfigure}{0.245\linewidth}
        \centering
        \includegraphics[width=\linewidth]{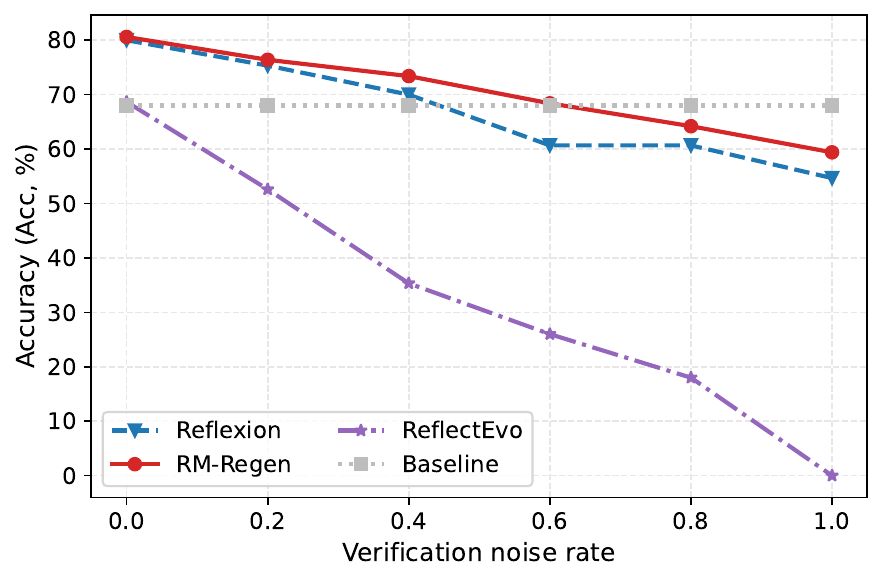}
        \caption{Letter}
        \label{fig:verification_noise_letter}
    \end{subfigure}
    \hfill
    \begin{subfigure}{0.245\linewidth}
        \centering
        \includegraphics[width=\linewidth]{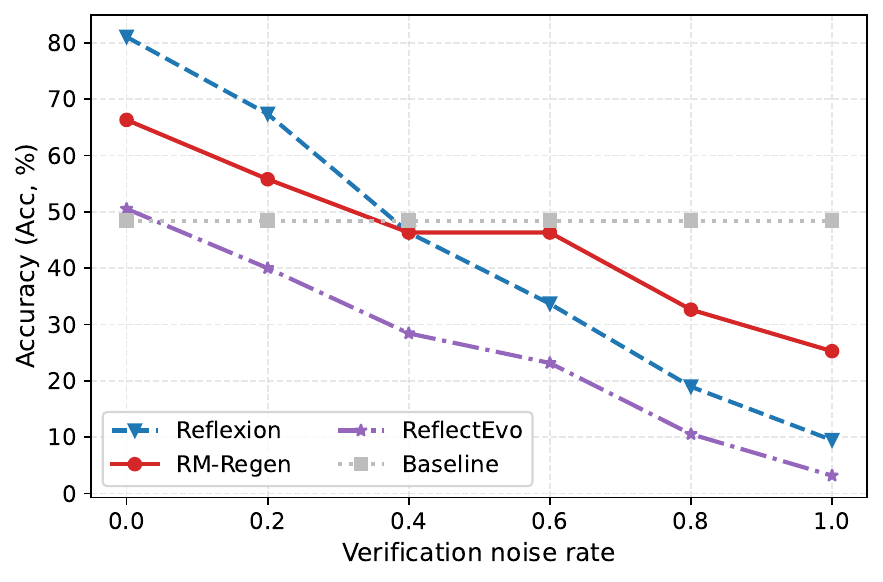}
        \caption{Legal Bench}
        \label{fig:verification_noise_legal}
    \end{subfigure}

    \caption{
   Performance robustness against verification noise. We compare our method with Reflexion and ReflectEvo on Llama-3.1-8B. The gray baseline denotes Few-shot CoT. The x-axis represents the noise rate, where $0$ denotes oracle verification and $1$ denotes verification is always incorrect.
    }
    \label{fig:verification_noise}
\end{figure*}

\subsection{RM-Guided Verification and Regeneration}
We further compare RM-Regen with several verification-based baselines. 
In the oracle setting, we include Reflexion, ReflectEvo, and Best-of-$N$ selection (upper bound), while in the non-oracle setting, we evaluate against Self-Refine, ProCo, and ST-CoT.

\subsubsection{With Oracle Verification}

Under the oracle verification setting, as shown in Table~\ref{tab:with oracle verification}, our method achieves the best overall performance. For GPT-3.5, RM-Regen obtains the highest average accuracy of 76.94\%, outperforming Reflexion (3 iterations) by 6.29\%, and Best-of-N (N=3) by 9.65\%. A similar trend holds for LLaMA~3.1-8B and Gemma~2-9B. RM-Regen achieves 75.85\% on LLaMA~3.1-8B, surpassing Reflexion (3 iterations) by 2.59\%, Best-of-N (N=3) by 6.77\%, and ReflectEvo by 11.11\%. On Gemma-2-9B, RM-Regen reaches 71.42\%, improving over Reflexion (3 iterations) by 4.02\% and Best-of-N (N=3) by 11.64\%. 
These gains indicate that our training-free regeneration can be more effective than iterative refinement and multi-generation strategies, and more generalizable than training-based approaches. Additional results for Reflexion with 2–5 iterations and Best-of-N (N=10) are given in Appendix~\ref{sec:full results reflexion}, where our method remains competitive.
Overall, these results indicate that given the same perfect verification signal, our method effectively transforms it into larger gains.

Since oracle verification is impractical in real-world settings, we analyze the impact of imperfect verification by progressively injecting noise into the oracle verifier. Verification noise is simulated by independently flipping each oracle label with a probability of the noise rate, as illustrated in Figure~\ref{fig:verification_noise}. It shows how different methods behave under varying levels of verification noise. As expected, the accuracy of all methods decreases monotonically as the verification noise rate increases. However, the rate of degradation differs substantially across methods. At zero noise, our method already achieves the highest accuracy, outperforming Reflexion and ReflectEvo. As the noise rate grows, the performance of both Reflexion and ReflectEvo drops sharply. In contrast, our method exhibits a much flatter decay curve and maintains a clear performance margin across all noise rates. Even when the verification signal becomes highly unreliable, our approach still preserves a substantial portion of its original accuracy. These trends indicate that our contrastive reflection mechanism is significantly more robust to incorrect verification signals than prior methods.

\subsubsection{Without Oracle Verification}

\begin{figure*}[htbp]
    \centering
    % 上边子图：MATH
    \begin{subfigure}{0.245\linewidth}
        \centering
        \includegraphics[width=\linewidth]{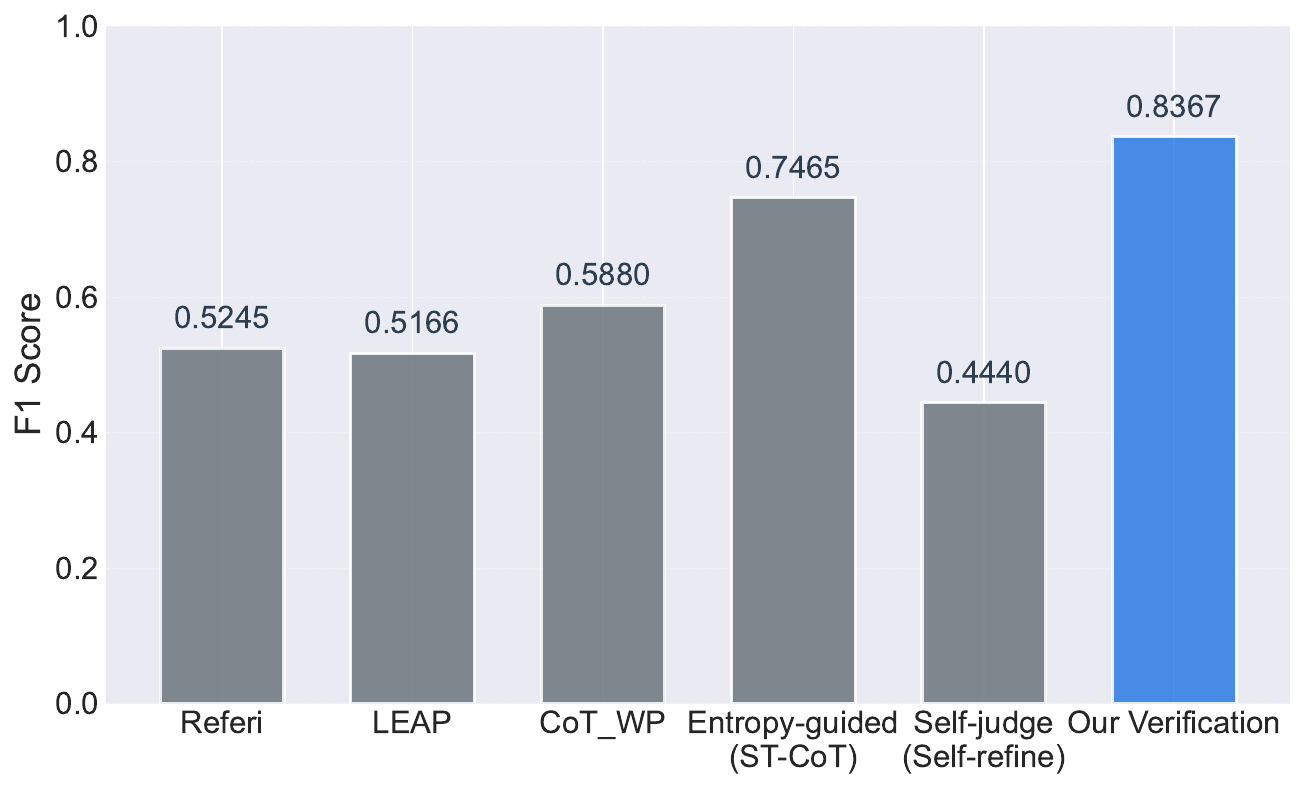}
        \caption{GSM\_Hard}
        \label{fig:verification_acc_math}
    \end{subfigure}
    \hfill
    % 下边子图：Bamboogle
    \begin{subfigure}{0.245\linewidth}
        \centering
        \includegraphics[width=\linewidth]{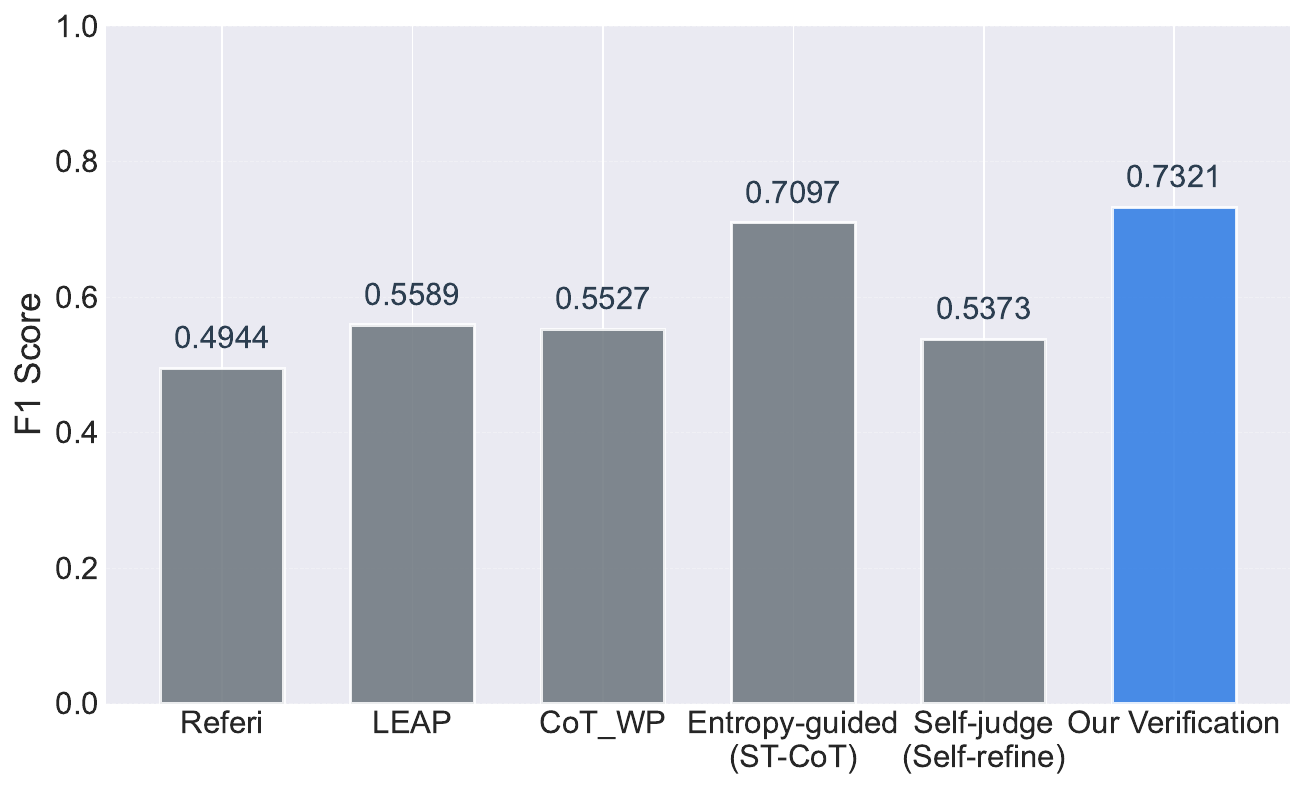}
        \caption{Bamboogle}
        \label{fig:verification_acc_bamboogle}
    \end{subfigure}
\hfill
    \begin{subfigure}{0.245\linewidth}
        \centering
        \includegraphics[width=\linewidth]{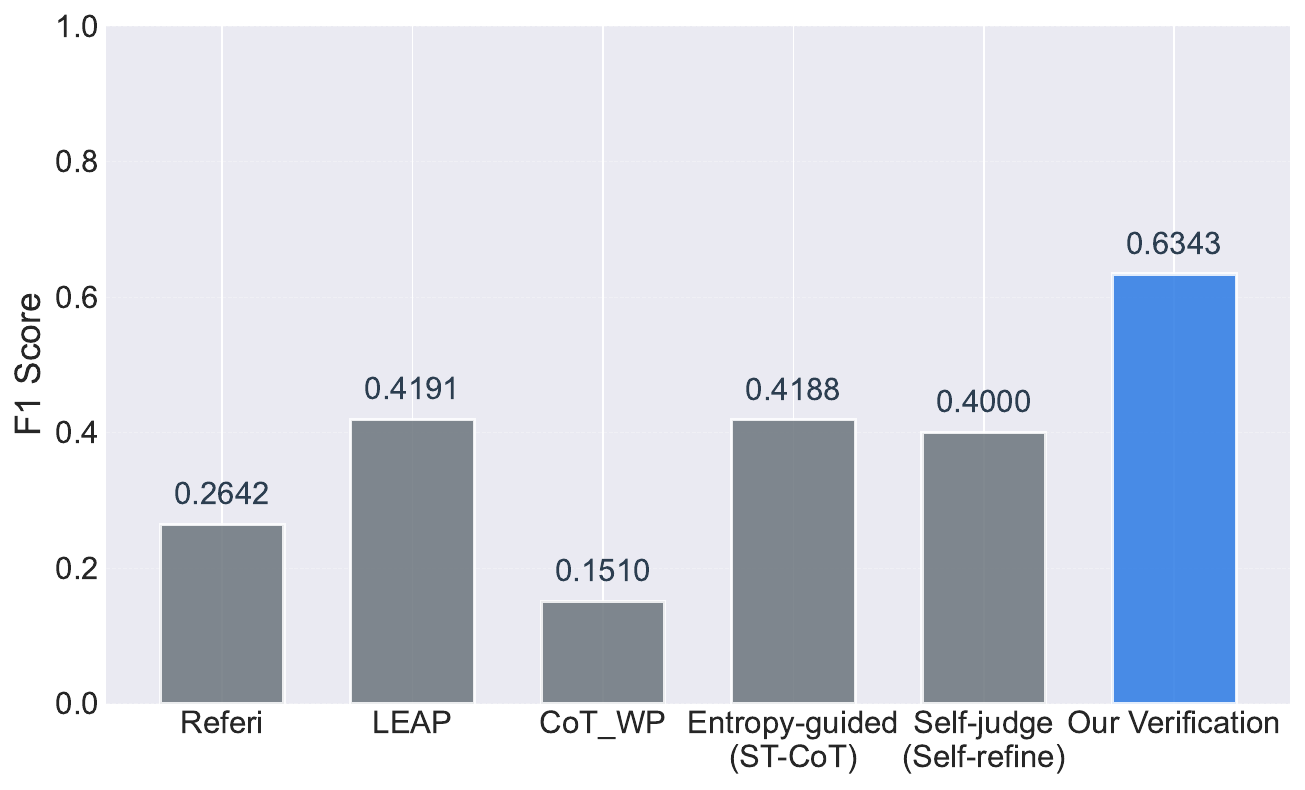}
        \caption{Coin}
        \label{fig:verification_noise_coin}
    \end{subfigure}
\hfill
    \begin{subfigure}{0.245\linewidth}
        \centering
        \includegraphics[width=\linewidth]{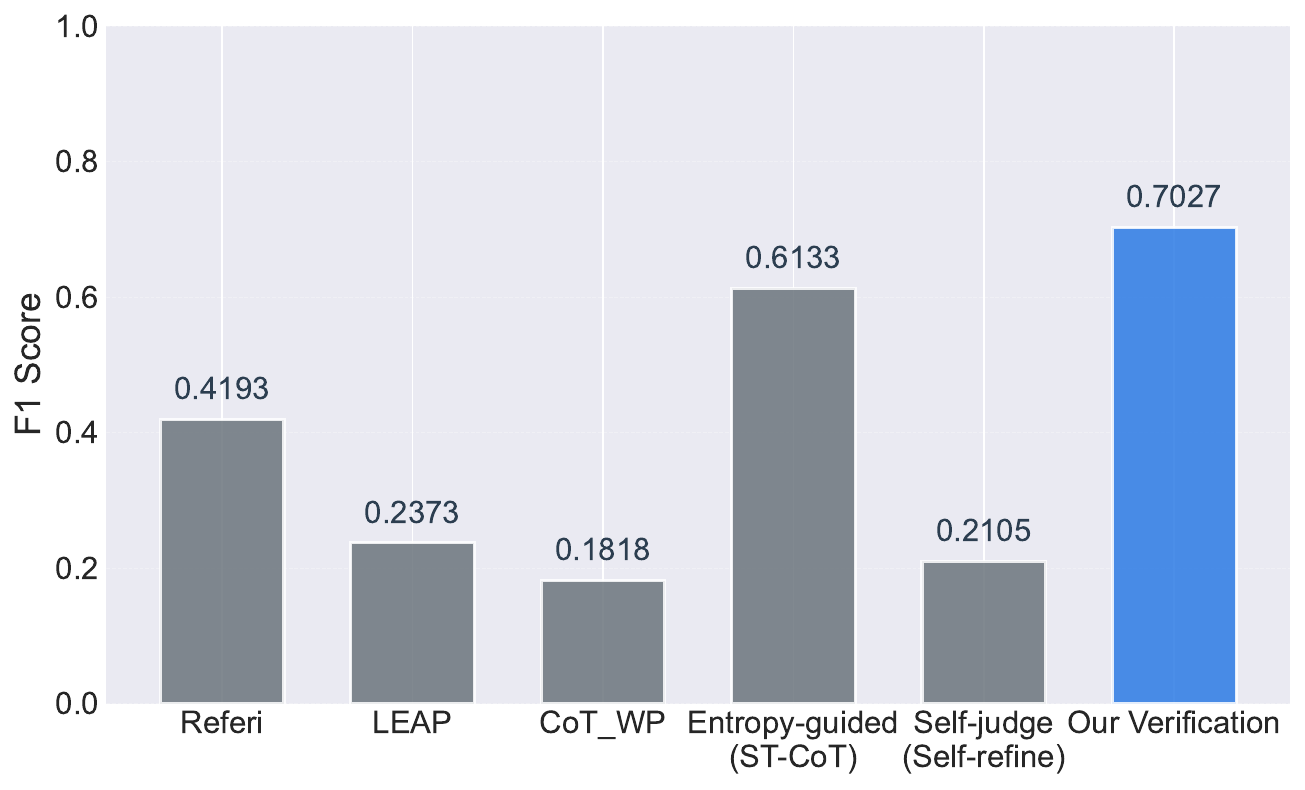}
        \caption{Legal Bench}
        \label{fig:verification_noise_legal}
    \end{subfigure}
    \caption{Verification F1 score of different verifiers on representative benchmarks using Llama-3.1-8B.}
    \label{fig:verification_acc}
\end{figure*}

\begin{table*}[t]
\centering
% 更紧凑的表格排版
\scriptsize
\setlength{\tabcolsep}{5pt}  % 列间距
\renewcommand{\arraystretch}{1.2} % 行距
\begin{tabularx}{0.83\textwidth}{cl *{10}{c}}
\toprule
& &\multicolumn{3}{c}{\textbf{Algorithmic}} & \multicolumn{2}{c}{\textbf{Reasoning}} & \multicolumn{2}{c}{\textbf{Symbolic}} & \multicolumn{2}{c}{\textbf{Domain}} & \multicolumn{1}{c}{\textbf{Overall}} \\
\cmidrule(lr){2-12}
\textbf{Model} & \textbf{Method} & GSM8K & GSM-H & MATH & SQA & Bamb & Coin & Letter & LB & HL & avg\%\\

\midrule
\multirow{5}{*}[4pt]{\rotatebox{90}{GPT-3.5}}
% &Self-Refine (2 iters)&73.40  & 33.80 & 48.20 & 44.00 & 40.00 & 67.25 & 40.00 & 53.68 & 59.00&52.17\\
&Self-Refine (3 iters)  &70.40  & 32.80 & 47.20 & 41.75 & 42.00 & 68.25 & 46.00 & 28.42 & 53.00&50.38\\
% &ProCo (2 iters) &80.06 &39.40 &54.60  & 56.75 &49.00  &73.00  &74.67 &48.42  &64.00 &60.48\\
&ProCo (3 iters) &80.80  & 39.60 & 52.20 & 60.00 & 43.00 & 75.25 & 74.00 & 41.05 & 65.00 & 60.55\\
% &ST CoT (2 iters)&82.20 &39.60 &54.80  & 62.00 & 51.00 & 75.50 &72.67 & 25.26 &70.00 &61.46\\
&ST CoT (3 iters)&80.20 & 39.80 & 53.60 & 62.00 & 59.00 & 75.00 & 72.67 & 23.16 & 69.00&61.02\\
&RM-Regen       &\textbf{83.60}&\textbf{64.00}&\textbf{56.80}&\textbf{69.75}&\textbf{59.00}&\textbf{78.50}&\textbf{77.33}&\textbf{60.00}&\textbf{71.00}&\textbf{69.87}\\
\midrule

\multirow{5}{*}[3pt]{\rotatebox{90}{Llama3.1-8B}}
% &Self-Refine (2 iters)&75.80 &23.40 & 54.80 &54.25  &47.00  &67.25  & 47.33 &46.32 &60.00 &53.84\\
&Self-Refine (3 iters)&75.60 &23.00 & 51.00 &53.75  &49.00  &66.00  & 40.00 &46.32 &65.00 &52.64\\
% &ProCo (2 iters)  &83.40 &31.00 &69.60&	59.75&	35.00&	42.00&	70.67&	40.00&	48.00& 56.61\\
&ProCo (3 iters)  &83.40 &31.20 & 69.60& 62.00& 39.00&44.00&71.33&38.95&48.00& 57.41\\
% &ST CoT (2 iters)&86.60 &31.40 &\textbf{72.20} &68.75  &48.00  &69.50  & 62.00 &54.74 &66.00 &64.23\\
&ST CoT (3 iters)&87.00 &31.80 & 69.80 &68.00  &48.00  &69.25  & 58.00 &55.79 & 68.00&63.68\\
&RM-Regen        &\textbf{87.20}&\textbf{34.80}&\textbf{71.40}&\textbf{73.00}&\textbf{52.00}&\textbf{77.75}&\textbf{79.33}&\textbf{56.84}&\textbf{73.00}&\textbf{68.05}\\

\midrule
\multirow{5}{*}[3pt]{\rotatebox{90}{Gemma2-9B}}
% &Self-Refine (2 iters)&  &   &  &   &   &   &   &   &\\
&Self-Refine (3 iters)&40.00 &32.80 &27.80 &28.00 &16.00 &54.25 &24.00 &41.05 &59.00 & 35.77\\
% &ProCo (2 iters)      &  &   &  &   &   &   &   &   &\\
&ProCo (3 iters)      &89.60 &48.60 &64.60 &63.00 &38.00 &50.75 &35.33 &44.21 &69.00& 60.87\\
% &ST CoT (2 iters)     &  &   &  &   &   &   &   &   &\\
&ST CoT (3 iters)     &88.00 &46.20 &59.40 &47.25 &46.00 &45.00 &33.33 &45.26 &66.00 & 56.17\\
&RM-Regen             &\textbf{89.80} &\textbf{49.40} &\textbf{66.20} &\textbf{66.00} &\textbf{48.00} &\textbf{66.00} &\textbf{37.86} &\textbf{51.58} &\textbf{71.00} &\textbf{64.84} \\
\bottomrule
\end{tabularx}
\caption{Accuracy on nine benchmarks without oracle verification (verification-and-rectification baselines vs. ours).}
\label{tab:without oracle verification}
\end{table*}

Under no oracle verification setting, we first evaluate our verification performance against existing verifiers: Referi~\cite{lee2025training}, LEAP~\cite{zhang2024context}, CoT\_WP~\cite{wang2024chain}, Entropy-guided (ST CoT), Self-judge (Self-Refine). Then, we evaluate the overall accuracy achieved by verification-guided improvement across different methods: Self-Refine, ProCo, and ST-CoT.

\textbf{Verifier Comparison}~
Figure~\ref{fig:verification_acc} compares the verification F1 scores of different verification strategies on GSM\_Hard, Bamboogle, Coin, and Legal. Our verifier attains an F1 score outperforming the strongest baseline, the entropy-guided verifier from ST-CoT. Other methods such as Referi, LEAP, CoT\_WP, and the self-judge verifier lag considerably behind.
The consistent gains across algorithmic, reasoning, symbolic and specific domain benchmarks suggest that our verifier generalizes well across tasks.
Overall, these results demonstrate that our verification scheme provides a more reliable signal for subsequent rectification.

\textbf{Overall Accuracy Comparison}~
Table~\ref{tab:without oracle verification} reports the performance of different verification-and-rectification methods without oracle verification signals. 
For GPT-3.5, our method achieves the highest overall average accuracy of 69.87\%, which represents a substantial 8.85\% improvement over the best baseline, ST CoT (3 iterations, 61.02\%). 
A similar pattern is observed for LLaMA~3.1-8B and Gemma~2-9B. Our approach attains an average accuracy of 68.05\% on Llama~3.1-8B, outperforming the strongest ST CoT variant (3 iterations, 63.68\%) by 4.37\%, and 64.84\% on Gemma~2-9B, exceeding the strongest ProCo variant (3 iterations, 60.87\%) by 3.97\%. 
These results suggest that under imperfect verification, reflection-style “explain-and-fix” updates are less reliable than direct regeneration with RM demonstrations, because noisy feedback introduces another generation step to propagate errors. Additional results for 2–5 iterations are provided in Appendix~\ref{sec:full results iteration}, showing consistent trends.
Overall, our method consistently improves reasoning accuracy across three base models, providing stronger gains than simply increasing refinement iterations in existing self-refinement methods.

\begin{figure}[htbp]
    \centering
    \includegraphics[width=0.9\linewidth]{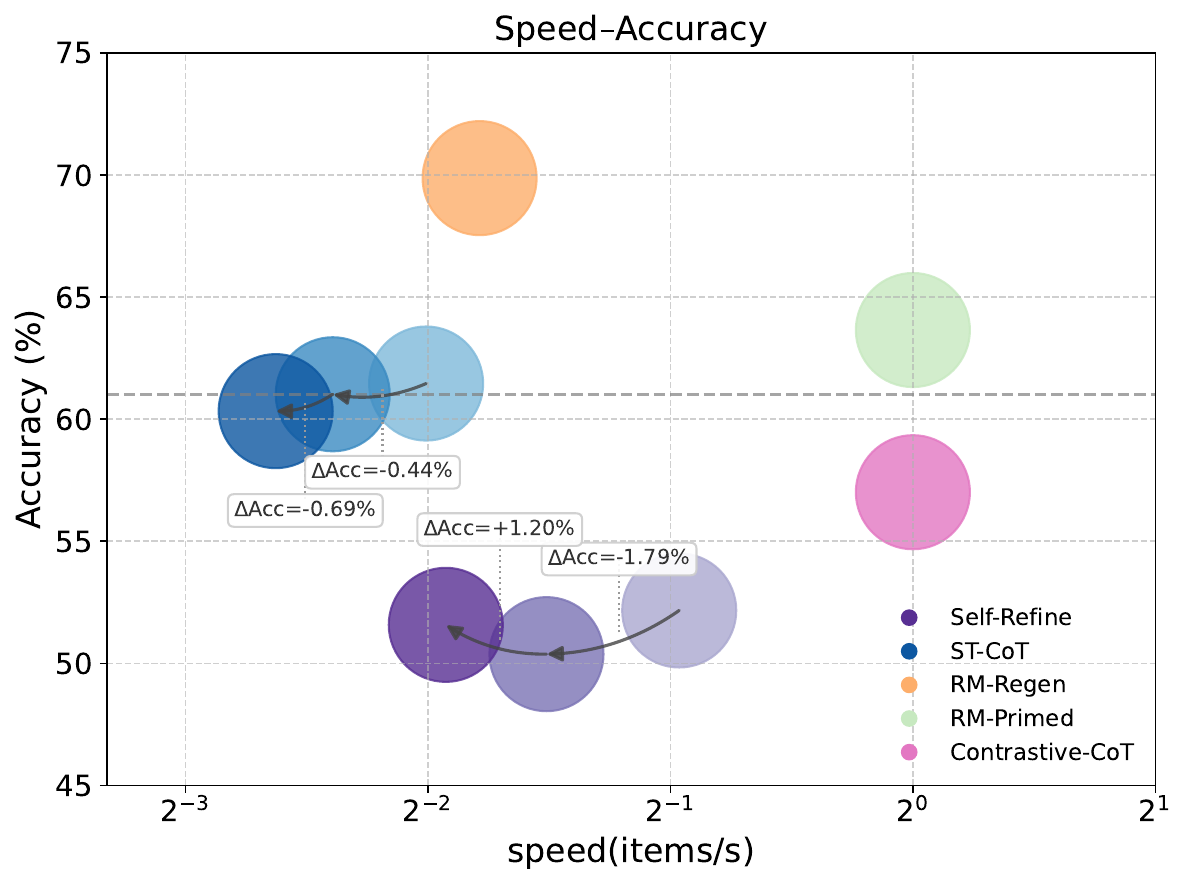} 
    \caption{Speed–accuracy trade-off of verification-guided improvement methods. Each bubble represents the average accuracy (y-axis) versus inference speed in items per second (x-axis, log-scale). Arrows connect configurations with different refinement iterations (2 iters - 4 iters) for the same method, and the attached labels report the resulting change in accuracy ($\Delta$Acc).}
    \label{fig:computation_complexity} 
\end{figure}

\textbf{Inference Speed-Accuracy Trade-Off}~
Figure~\ref{fig:computation_complexity} summarizes the inference speed–accuracy trade-off of verification-guided improvement methods. Iterative baselines, Self-Refine and ST-CoT, fall in the lower-left region, showing slower inference with only moderate accuracy. As the number of refinement iterations increases, inference speed drops while the accuracy gains are small or even negative. In contrast, our two strategies, RM-Primed and RM-Regen, lie in the upper-right region, achieving higher accuracy at higher speed. Contrastive-CoT attains slightly better speed than Self-Refine and ST-CoT, but its accuracy remains clearly below ours. 
Although RM is constructed once offline using teacher calls, this cost is amortized over large-scale inference, making the framework well suited for high-throughput serving scenarios.

\subsection{Teacher Model Dependency Analysis}
\label{sec:teacher model dependency}
\begin{table}[htbp]
\centering
\scriptsize
\begin{tabularx}{0.84\columnwidth}{l *3{c}}
\toprule
Teacher Model  & \textbf{GSM-Hard} & \textbf{Bamboogle} & \textbf{Coin}  \\
\midrule
\textcolor{gray!70}{\textbf{No Teacher (CoT)}}     &\textcolor{gray!80}{31.80} &\textcolor{gray!80}{51.00} &\textcolor{gray!80}{65.75}\\
\textbf{GPT-5-mini}     &\textbf{41.00} &\textbf{63.00} &\textbf{85.75}\\
\textbf{GPT-3.5}        &37.60 &60.00 &76.50\\
\textbf{Llama~3.1-8B}   &39.00 &60.00 &74.50\\
\textbf{Gemma~2-9B}     &39.60 &58.00 &77.75\\
\bottomrule
\end{tabularx}
\caption{Teacher model dependency analysis. Performance evaluated with base model Llama~3.1-8B across different teachers.}
\label{tab:teacher denpendency}
\end{table}
To study the impact of teacher model, we construct RMs using teachers of varying capabilities and evaluate performance with based model Llama~3.1-8B. In Table~\ref{tab:teacher denpendency}, GPT-5-mini, the strongest teacher, achieves the highest accuracy across datasets, while weaker teachers such as GPT-3.5, LLaMA-3.1-8B, and Gemma-2-9B cause slight drops, but still improve over few-shot CoT without a teacher.
These results indicate that stronger teachers provide better signals for RM construction, but overall performance does not critically depend on any specific teacher. RMs distilled from moderately capable models remain effective, capturing robust and transferable reasoning patterns.
Further ablation studies, as well as analyses of RM size, regenerate rate, flip, and out-of-distribution performance, are provided in Appendices~\ref{sec:ablation},~\ref{sec:RM size},~\ref{sec:wrong-to-correct},~\ref{sec:out-of-domain}.

 \section{Related Work}
% \subsection{Structured Reasoning Exploration}
% CoT prompting \cite{wei2022chain} elicits intermediate steps to improve LLM reasoning, inspiring methods that leverage explicit reasoning. Building on this, \cite{wang2023self} introduced self-consistency decoding, which—rather than greedy decoding—samples diverse reasoning paths and aggregates them into a single answer. Subsequently, more advanced frameworks such as Tree-of-Thought (ToT) \cite{yao2023tree} and Algorithm-of-Thought (AoT) \cite{sel2024algorithm} were developed. ToT explores the reasoning space step by step; when decomposition is difficult, it reduces to generating multiple full solutions and selecting the best. By contrast, AoT does not trigger exploration at every step; instead, it teaches exploration via in-context examples so the entire search can be executed in a single run. While this reduces the number of model invocations, single-run exploration demands stronger models and was validated primarily on GPT-4. Forest-of-Thought (FoT) \cite{bi2025forest} further broadens search structures. However, these methods primarily improve performance by expanding structured exploration while largely overlooking verification and rectification—i.e., how to assess and efficiently correct reasoning errors once a candidate trajectory is produced.
 
\textbf{ICL Demonstrations Improves Reasoning}~
Well-designed few-shot demonstrations can boost LLM performance. Auto-CoT \cite{zhang2023automatic} automatically generates CoT examples and, with proper selection, can outperform hand-crafted CoT. Building on this, contrastive CoT \cite{chia2023contrastive} and related methods \cite{mo2024cicl,gao2024customizing} present, via in-context prompts, paired correct/incorrect answers augmented with analyses so that models learn which mistakes to avoid. \citet{gu-etal-2025-toward} further extends this by retrieving contrastive experience from memory to support reasoning. 
However, these methods typically either omit explicit output-level verification or lack structured reflection--correction pairs for error repair.

\noindent \textbf{Self-Reflection and Self-Refinement}~
Leveraging LLMs’ self-assessment and refinement abilities has emerged as a promising direction. Several training-free methods have been proposed: Self-Refine \cite{madaan2023self} iteratively refines outputs via self-assessment; Reflexion \cite{shinn2023reflexion} lets LLMs verbally reflect on feedback and store reflections in memory buffer to guide future trials; \citet{wu2025rethinking} shows that iteratively reusing model-generated information can reduce prediction uncertainty in a self-training–style framework. 
For small-scale LLMs (SLLMs), fine-tuning–based approaches have also been explored. ReflectEvo \cite{li2025reflectevo} builds a reflection learning dataset and improves SLLM reasoning via SFT and DPO. However, these methods typically rely on ground-truth verification or iterative rectification, which may accumulate errors and become trapped in faulty reasoning.
Recent work has explored reinforcement learning to strengthen the “verify–improve” loop. Methods such as TTRL \cite{zuo2025ttrl} and RESTRAIN \cite{yu2025restrain} apply test-time optimization or self-penalized RL to stabilize self-correction without gold labels. Evolution-based selection \cite{zhou2025on} similarly improves reasoning through intensive sampling. In contrast, our RM-based framework is training-free and avoids online policy updates or excessive sampling. Our method enables efficient error correction through lightweight retrieval and prompting, making it suitable for deployment scenarios where real-time RL is computationally expensive or unstable.

\section{Conclusion}
This paper proposes a training-free regeneration paradigm in which Reflection Memory (RM) provides contextual guidance to correct reasoning errors and regeneration from scratch enables break out of faulty reasoning loops. Under oracle verification, our method achieves the best performance and remains the most robust with noisy verification. Without oracle verification, RM–guided self-verification attains the highest F1 score, reliably triggering regeneration. Overall, this paradigm delivers significant gains in LLM reasoning performance while maintaining low computational cost.

\section{Limitations}
Although our training-free regeneration paradigm achieves strong performance across diverse tasks, it still has several limitations that need further investigation.
First, the quality of RM is highly dependent on the teacher model. Without a sufficiently powerful teacher to generate reliable reflections, summarize principles, and provide accurate rectifications, the effectiveness of RM can degrade substantially.
Although our verifier attains high recall, it is still imperfect, and some incorrect responses remain undetected. These missed errors are a bottleneck that limits the overall performance ceiling.
Future work could improve RM quality through multi-teacher aggregation or distillation, and develop more reliable RM-aware verifiers, such as lightweight learned or ensemble-style models that leverage logged missed errors for offline calibration. These directions may further reduce undetected mistakes while largely preserving the single-pass regeneration paradigm.

\section{Ethical Considerations}
This work aims to improve LLMs' performance through response verification and rectification. 
However, imperfect verification can still miss errors or incorrectly modify correct answers, potentially leading to overconfidence or unnecessary rewrites. 
In addition, the same verification and rewriting mechanisms could be misused to produce more persuasive but incorrect content; therefore, we recommend cautious use in safety-critical settings and emphasize the need for human oversight.
We do not collect new personal data, and all experiments are conducted on publicly available benchmarks.

% \section*{Acknowledgments}
% xxxxxxxx

% Bibliography entries for the entire Anthology, followed by custom entries
%\bibliography{anthology,custom}
% Custom bibliography entries only
\bibliography{custom}

\clearpage
\appendix

\begin{table*}[!t]
\centering
% 更紧凑的表格排版
\scriptsize
\setlength{\tabcolsep}{4pt}  % 列间距
\renewcommand{\arraystretch}{1.1} % 行距
\begin{tabularx}{0.88\textwidth}{l *{9}{c}}
\toprule
& \multicolumn{3}{c}{\textbf{Algorithmic}} & \multicolumn{2}{c}{\textbf{Reasoning}} & \multicolumn{2}{c}{\textbf{Symbolic}} & \multicolumn{2}{c}{\textbf{Domain}} \\
\cmidrule(lr){2-10}
\textbf{Dataset} & GSM8K\footnotemark[1] & GSM-Hard\footnotemark[2] & MATH\footnotemark[3] & StrategyQA\footnotemark[4] & Bamboogle\footnotemark[5] & Coin\footnotemark[6] & Letter\footnotemark[7] & LB\footnotemark[8] & Headline\footnotemark[9] \\
\midrule
\textbf{Train split}& 800 & 800 & 800 & 800 & 10  & 100 & 10  & 5  & 100 \\
\textbf{Test split} & 500 & 500 & 500 & 400 & 100 & 400 & 140 & 95 & 100 \\
\textbf{Answer type} & Numeric & Numeric & Numeric & Boolean & Factoid & Boolean & Letter & Multiple-choice & Multiple-choice \\
\bottomrule
\end{tabularx}
\caption{Dataset splits used in our experiments. For each benchmark, we report the number of train and test examples used for reflection-memory construction and evaluation. We randomly subsample train/test examples from each benchmark using a fixed random seed, ensuring the same subset across all methods.}
\label{tab:data splits}

\vspace{4em}

\centering
% 更紧凑的表格排版
\scriptsize
\setlength{\tabcolsep}{4pt}  % 列间距
\renewcommand{\arraystretch}{1.1} % 行距
\begin{tabularx}{0.71\textwidth}{l *{9}{c}}
\toprule
& \multicolumn{3}{c}{\textbf{Algorithmic}} & \multicolumn{2}{c}{\textbf{Reasoning}} & \multicolumn{2}{c}{\textbf{Symbolic}} & \multicolumn{2}{c}{\textbf{Domain}} \\
\cmidrule(lr){2-10}
\textbf{Dataset} & GSM8K & GSM-Hard & MATH & StrategyQA & Bamboogle & Coin & Letter & LB & Headline \\
\midrule
\textbf{GPT-5mini} &94.00  &75.00 &97.80  &80.75  &66.00   & 79.50  & 96.67    & 84.21    &79.00 \\
\bottomrule
\end{tabularx}
\caption{Performance of teacher model on nine benchmarks.}
\label{tab:teacher}
\end{table*}

\section{Implementation Details}
\label{sec: implementation detail}

\subsection{Dataset Details and Splits}
The evaluation covers nine benchmarks spanning algorithmic, reasoning, symbolic, and domain-specific tasks, providing a diverse test cases for assessing both general problem-solving ability and specialized-domain reliability. These benchmarks vary in input formats, reasoning requirements, and answer types, allowing us to examine the robustness of each method across heterogeneous settings rather than on a single task family. The corresponding dataset statistics, including the number of examples and the evaluation splits used in our experiments, are summarized in Table~\ref{tab:data splits}.
All datasets are obtained from publicly available sources; license and usage terms follow the corresponding dataset cards and original releases.

\footnotetext[1] {\url{https://huggingface.co/datasets/gsm8k}}
\footnotetext[2]{\url{https://huggingface.co/datasets/reasoning-machines/gsm-hard}}
\footnotetext[3]{\url{https://huggingface.co/datasets/nlile/hendrycks-MATH-benchmark}}
\footnotetext[4]{\url{https://huggingface.co/datasets/tasksource/strategy-qa}}
\footnotetext[5]{\url{https://huggingface.co/datasets/chiayewken/bamboogle}}
\footnotetext[6]{\url{https://huggingface.co/datasets/skrishna/coin_flip}}
\footnotetext[7]{\url{https://huggingface.co/datasets/ChilleD/LastLetterConcat}}
\footnotetext[8]{\url{https://huggingface.co/datasets/nguha/legalbench}}
\footnotetext[9]{\url{https://huggingface.co/datasets/SaguaroCapital/sentiment-analysis-in-commodity-market-gold}}

\subsection{Inference and Hyperparameter Settings}
\label{subsec: hyperparameter setting}
We evaluate both proprietary and open-source LLMs. Specifically, we query GPT-3.5-0125 and GPT-5-mini-2025-08-07 via the OpenAI API, while LLaMA~3.1-8B-Instruct, Gemma~2-9B-it is deployed locally using vLLM on an NVIDIA A6000 GPU for efficient batched inference.

\noindent
\textbf{Inference-time decoding details}~ We use a unified decoding configuration across all benchmarks and methods to ensure a controlled comparison. When multiple generations are required (e.g., sampling-based verification), we adopt sampling with temperature = 0.3 and top\_p = 0.7, and cap the maximum generation length at 1024 tokens ({max\_tokens} = 1024). For verification-based components, we draw three independent samples per query under the same sampling settings and feed the resulting verification outcomes into the corresponding verification pipeline. 
When only a single generation is needed, we switch to deterministic decoding to improve stability and reproducibility, using temperature = 0 and top\_p = 1 (with the same \texttt{max\_tokens} limit). All other inference parameters are kept at their default values for the OpenAI API and vLLM, respectively.

\noindent
\textbf{Retrieval and reranking hyperparameters}~ For the retrieval-then-reranking pipeline, we select the number of retrieved candidates $n$ and the number of reranked memories $k$ based on a trade-off between retrieval coverage and inference efficiency. A larger $n$ improves the likelihood of recalling relevant memories at the cost of additional computation, while a smaller $k$ helps reduce noise and stabilize in-context learning. In our experiments, we fix $n = 10$ and $k = 3$ across all benchmarks, as these values provided stable performance. We found that moderate variations around these settings lead to similar trends, indicating that our method is not overly sensitive to the exact choice of n and k.

\subsection{Performance of the Teacher Model}
We report the teacher model’s Few-shot CoT performance in Table~\ref{tab:teacher}. Overall, the teacher achieves consistently strong accuracy across the evaluated benchmarks, indicating that it can serve as a reliable source for generating RM (including reflection, principle, and rectification) in our pipeline. In particular, its high correctness rate suggests that the produced reflections are more likely to identify the true failure modes of incorrect responses, and that the rectified demonstrations provide accurate, task-relevant reasoning patterns. This is important because the quality of both reflection and correction is bounded by the teacher’s competence; a sufficiently strong teacher reduces noisy feedback and improves the usefulness of the resulting reflection memories.

\subsection{Reproduction Details of Baselines}

\setcounter{footnote}{9}

\noindent
\textbf{Training-based baselines.}~
We treat ReflectEvo\footnote{\url{https://github.com/bigai-nlco/ReflectEvo}} as a \emph{training-based} baseline.
ReflectEvo considers four training configurations:
(i) one-stage training of both self-reflection and self-correction on $D^{+}$; 
(ii) two-stage training of self-reflection and self-correction on $D^{+}$;
(iii) self-reflection only on $D^{\pm}$; and 
(iv) self-reflection only on $D^{\mathrm{pref}}$.
In our experiments, we reproduce the two-stage w/ $D^{+}$ setting, which achieves the best results in the LogiQA task in the original paper.
For algorithmic tasks (GSM8K, GSM-Hard, and MATH), we fine-tune Llama-3.1-8B using the \texttt{math} training data released by ReflectEvo.
For the remaining benchmarks, we fine-tune Llama-3.1-8B on the \texttt{logiqa} training data to assess the generalizability of ReflectEvo beyond its original training domain.

\noindent
\textbf{Training-free prompt-based baselines.}~
We treat Self-Refine\footnote{\url{https://github.com/madaan/self-refine}}, Reflexion\footnote{\url{https://github.com/noahshinn/reflexion}}, ST-CoT\footnote{\url{https://github.com/zongqianwu/ST-COT}}, ProCo\footnote{\url{https://github.com/wzy6642/ProCo}}, and Contrastive-CoT\footnote{\url{https://github.com/DAMO-NLP-SG/contrastive-cot}} as \emph{training-free prompt-based} baselines, since they do not update model parameters and differ only in their prompting and inference procedures.
For each baseline, we carefully reproduce the core algorithm in our unified evaluation framework: we read the original paper and public implementation, extract the key inference steps, and re-implement them on top of our codebase while keeping the underlying LLM, decoding settings, and evaluation splits identical across methods.
We follow the original prompt templates as closely as possible and adapt only the task-specific content.
For each benchmark, we construct a fixed set of few-shot demonstrations from the training split; unless otherwise specified, all methods use the same set and number of in-context examples on a given dataset.
For iterative baselines, we evaluate multiple numbers of refinement iterations, and report the corresponding results in Sec.~\ref{sec:Experiment}.

\section{Full Benchmark Results}
\label{sec:full results}

\subsection{Direct Performance Boost}
\label{sec:full results direct boost}

\begin{table*}[!t]
\centering
% 更紧凑的表格排版
\scriptsize
\setlength{\tabcolsep}{5pt}  % 列间距
\renewcommand{\arraystretch}{1} % 行距
\begin{tabularx}{0.9\textwidth}{l *{10}{c}}
\toprule
& \multicolumn{3}{c}{\textbf{Algorithmic}} & \multicolumn{2}{c}{\textbf{Reasoning}} & \multicolumn{2}{c}{\textbf{Symbolic}} & \multicolumn{2}{c}{\textbf{Domain}} & \multicolumn{1}{c}{\textbf{Overall}} \\
\cmidrule(lr){2-11}
\textbf{Method} & GSM8K & GSM-Hard & MATH & StrategyQA & Bamboogle & Coin & Letter & LB & Headline & avg\%\\
\midrule

\multicolumn{11}{l}{\textbf{GPT-3.5}} \\
\midrule
Few-shot CoT           &79.25  & 41.25         & 51.60 & 61.50         & 55.00         & 74.25         & 72.00 & 44.21         & 70.00 &61.15\\
Contrastive CoT (w/o r) &77.20  & 39.40         & 47.00 & 60.75         & 49.00         & 59.00         & 75.33 & 35.75         & 72.00 &57.01\\
Contrastive CoT (w r)   &78.40  & 38.00         & 52.20 & 63.00         &50.00          & 69.25         & 73.33 & 55.79         & 71.00 &60.33\\
RM-Primed ($\mathcal{R}^{+}$)&81.50  & 42.50         &\textbf{55.00}& 65.25         & 59.00         & 78.00         &\textbf{84.00}& 64.21         & 70.00 &64.99\\
RM-Primed                   & \textbf{81.80}& \textbf{44.60}& 54.80 & \textbf{68.75}& \textbf{62.00}& \textbf{79.75}& 76.00 & \textbf{67.37}& \textbf{72.00} & \textbf{66.01}\\
\midrule

\multicolumn{11}{l}{\textbf{Llama 3.1-8B}} \\
\midrule
Few-shot CoT           & 84.80 & 31.80 &\textbf{71.40}& 69.25         & 51.00          & 65.75         & 68.00         & 48.42         & 67.00 &64.19\\
Contrastive CoT (w/o r) & 76.2  & 33.20 & 66.00 & 69.25         & 25.00          & 60.75         & 61.33         & 57.89         & 66.00 &59.56\\
Contrastive CoT (w r)   &76.40  & 33.00 & 67.60 & 70.25         & 29.00          & 66.25         & 62.67         & 53.68         & 73.00 &61.13\\
RM-Primed ($\mathcal{R}^{+}$)& 84.60 &\textbf{35.80}& 64.40 & 70.25         & 51.00          & 78.75         & 72.00         & 62.11         & 72.00 &65.94\\
RM-Primed                   &\textbf{85.20}& 34.20 & 70.00 &\textbf{73.25} & \textbf{54.00} & \textbf{80.75}& \textbf{81.33}& \textbf{64.21}& \textbf{73.00}&\textbf{68.23}\\
\bottomrule
\end{tabularx}
\caption{Full per-dataset accuracy of different direct-boost methods on nine reasoning benchmarks.}
\label{tab:full Direct Performance Boost}

\vspace{4em}

\centering
% 更紧凑的表格排版
\scriptsize
\setlength{\tabcolsep}{5pt}  % 列间距
\renewcommand{\arraystretch}{1} % 行距
\begin{tabularx}{0.9\textwidth}{l *{10}{c}}
\toprule
& \multicolumn{3}{c}{\textbf{Algorithmic}} & \multicolumn{2}{c}{\textbf{Reasoning}} & \multicolumn{2}{c}{\textbf{Symbolic}} & \multicolumn{2}{c}{\textbf{Domain}} & \multicolumn{1}{c}{\textbf{Overall}} \\
\cmidrule(lr){2-11}
\textbf{Method} & GSM8K & GSM-Hard & MATH & StrategyQA & Bamboogle & Coin & Letter & LB & Headline & avg\%\\
\midrule

\multicolumn{11}{l}{\textbf{GPT-3.5}} \\
\midrule
Reflexion (2 iters)      & 84.00         & 44.00         & 59.00         & 67.75         & 63.00         & 87.25 & 76.00         & 62.11 & 72.00&67.87\\
Reflexion (3 iters)      & 85.60         & 45.20         & 62.25         & 71.00         & 66.00         &{88.25}& 78.67         &{85.26} & 72.00&70.65\\
Reflexion (4 iters) &88.20&	48.80&	62.80&	73.00&	\textbf{76.00}&	90.50&	78.67&	91.58&	\textbf{74.00}& 73.15\\
Reflexion (5 iters) &86.20&	47.40&	64.20&	75.50&	70.00&	\textbf{91.50}&	81.33&	\textbf{92.63}&	73.00& 73.22\\
RM-Regen ($\mathcal{R}^{+}$) & 88.20         & 64.20         &\textbf{64.60} &\textbf{78.75} & 72.00         & 85.25 & 84.67         & 71.58 & 71.00&75.74\\
RM-Regen                    & \textbf{90.20}& \textbf{68.60}& 63.40         & 77.25         &{73.00} & 85.75 & \textbf{87.33}& 75.79 &{73.00}&\textbf{76.94}\\
\midrule

\multicolumn{11}{l}{\textbf{Llama 3.1-8B}} \\
\midrule
Reflexion (2 iters)      & 88.60 & 34.40         & 75.40         & 80.75         & 59.00         & 75.50         & 76.67         & 78.95         & 68.00&70.46\\
Reflexion (3 iters)      & 89.60 & 36.20         & 77.80         &{85.00} & 60.00         & 81.25         & 80.67         &{81.05}         & 70.00&73.26\\
Reflexion (4 iters) &89.40&	37.60&	78.80&\textbf{90.25}&	60.00&	82.00&	82.00&	82.11&	73.00& 74.75\\
Reflexion (5 iters) &89.80&	38.20&{80.60}&\textbf{90.25}&	60.00&	82.75&	82.00&	\textbf{87.37}&	71.00& 75.48\\
RflecEvo               & 85.40 & 32.00         & 73.60         & 71.25         & 53.00         & 67.75         & 68.67         & 50.53         & 62.00&64.74\\
Best-of-N (N=3) & 89.60&	34.20&	76.80&	77.50&	53.00&	72.00&	70.67&	64.21&	75.00& 69.08\\
Best-of-N (N=10) & 91.60&	36.00&	\textbf{81.00}&	81.25&	52.00&	72.50&	70.67&	72.63&	73.00& 71.33\\
RM-Regen ($\mathcal{R}^{+}$) & 91.80 & 38.20         & 79.40         & 81.50         & 60.00         & 85.00         & 86.67         & 65.26         & 74.00&74.28\\
RM-Regen             &\textbf{92.60}&\textbf{41.00} &{80.60} &{83.25}        &\textbf{63.00} & \textbf{85.75}&\textbf{89.33} & {66.32}&\textbf{75.00}&\textbf{75.85}\\
\bottomrule
\end{tabularx}
\caption{Accuracy on nine benchmarks with oracle verification. This table extends by additionally reporting Reflexion with 4 and 5 iterations, and best-of-N with N=10.}
\label{tab:full with oracle verification}

\end{table*}

The complete results of different direct performance boost methods are reported in Table~\ref{tab:full Direct Performance Boost}, which extends the compact summary in Table~\ref{tab:direct-boost-compact}. Specifically, we present per-dataset accuracies on all nine benchmarks for both GPT-3.5 and LLaMA~3.1-8B, enabling a fine-grained comparison beyond the averaged scores in the main text. 

Overall, our method outperforms other methods on most tasks and across both tested LLMs; the only exception is MATH on Llama-3.1-8B, where it slightly underperforms the few-shot CoT baseline.
We conjecture that this gap may stem from MATH’s high sensitivity to precise step-by-step calculations and strict final-answer formats; in such cases, adding RM to already solvable cases may reduce the benefits.

\subsection{With Oracle Verification}
\label{sec:full results reflexion}
Table~\ref{tab:full with oracle verification} summarizes the full per-dataset results under the oracle verification setting, complementing Table~\ref{tab:with oracle verification} in the main text. In addition to our methods, we report Reflexion with 2--5 iterations for both GPT-3.5 and Llama~3.1-8B, as well as Best-of-$N$ selection with larger candidate sets for Llama~3.1-8B. 

Several observations can be made. First, for Reflexion, increasing the number of iterations generally leads to performance gains across most datasets, indicating that iterative refinement can be beneficial when verification is oracle. However, these gains are not uniform: improvements tend to saturate or fluctuate on certain benchmarks.
Second, Best-of-$N$ selection exhibits only limited improvements as $N$ increases. While larger candidate pools occasionally yield higher accuracy, the overall gains remain modest, implying that naive candidate expansion alone is insufficient to consistently boost performance, even under oracle verification. Finally, our method consistently outperforms iterative Reflexion variants and Best-of-$N$ baselines in terms of average accuracy, indicating that memory-guided regeneration provide a more effective improvement mechanism than increasing refinement iterations or candidate counts.

\subsection{Without Oracle Verification}
\label{sec:full results iteration}

\begin{table*}[!t]
\centering
% 更紧凑的表格排版
\scriptsize
\setlength{\tabcolsep}{5pt}  % 列间距
\renewcommand{\arraystretch}{1} % 行距
\begin{tabularx}{0.9\textwidth}{l *{10}{c}}
\toprule
& \multicolumn{3}{c}{\textbf{Algorithmic}} & \multicolumn{2}{c}{\textbf{Reasoning}} & \multicolumn{2}{c}{\textbf{Symbolic}} & \multicolumn{2}{c}{\textbf{Domain}} & \multicolumn{1}{c}{\textbf{Overall}} \\
\cmidrule(lr){2-11}
\textbf{Method} & GSM8K & GSM-Hard & MATH & StrategyQA & Bamboogle & Coin & Letter & LB & Headline & avg\%\\
\midrule

\multicolumn{11}{l}{\textbf{GPT-3.5}} \\
\midrule
Self-Refine (2 iters)&73.40  & 33.80 & 48.20 & 44.00 & 40.00 & 67.25 & 40.00 & 53.68 & 59.00&52.17\\
Self-Refine (3 iters)  &70.40  & 32.80 & 47.20 & 41.75 & 42.00 & 68.25 & 46.00 & 28.42 & 53.00&50.38\\
Self-Refine (4 iters)&70.60  & 34.60 & 48.80 & 44.75 & 43.00 & 69.75 & 41.33 & 22.11 & 62.00 &51.58\\
ProCo (2 iters) &80.06 &39.40 &54.60  & 56.75 &49.00  &73.00  &74.67 &48.42  &64.00 &60.48\\
ProCo (3 iters) &80.80  & 39.60 & 52.20 & 60.00 & 43.00 & 75.25 & 74.00 & 41.05 & 65.00 & 60.55\\
ProCo (4 iters) &81.80   &39.40   &52.08	&55.75	&50.00	&73.75	&68.67	&42.11	&69.00& 60.11\\
ST CoT (2 iters)&82.20 &39.60  &54.80  & 62.00 & 51.00 & 75.50 &72.67 & 25.26 &70.00 &61.46\\
ST CoT (3 iters)&80.20 & 39.80 & 53.60 & 62.00 & 59.00 & 75.00 & 72.67 & 23.16 & 69.00&61.02\\
ST CoT (4 iters)&80.60 & 39.60 & 53.00 & 60.25 & 48.00 & 74.50 & 72.67 & 25.26 & 70.00 &60.33\\
RM-Regen        &\textbf{83.60}&\textbf{64.00}&\textbf{56.80}&\textbf{69.75}&\textbf{59.00}&\textbf{78.50}&\textbf{77.33}&\textbf{60.00}&\textbf{71.00}&\textbf{69.87}\\
\midrule

\multicolumn{11}{l}{\textbf{Llama 3.1-8B}} \\
\midrule
Self-Refine (2 iters)&75.80 &23.40 & 54.80 &54.25  &47.00  &67.25  & 47.33 &46.32 &60.00 &53.84\\
Self-Refine (3 iters)&75.60 &23.00 & 51.00 &53.75  &49.00  &66.00  & 40.00 &46.32 &65.00 &52.64\\
Self-Refine (4 iters)&75.60 &22.80 & 53.20 &55.00  &45.00  &64.25  & 41.33 &38.95 &63.00 &52.53\\
Self-Refine (5 iters)&75.40&21.40&	49.60&	51.25&	46.00&	66.00&	30.67&	44.21&	62.00& 50.89\\
ProCo (2 iters) &83.40&	31.00&	69.60&	59.75&	35.00&	42.00&	70.67&	40.00&	48.00& 56.61\\
ProCo (3 iters) & 83.40 &31.20 & 69.60& 62.00& 39.00&44.00&71.33&38.95&48.00& 57.41\\
ProCo (4 iters) &80.40&	31.80&	68.20&	64.75&	48.00&	48.00&	64.67&	44.21&	48.00&57.85\\
ProCo (5 iters) & 82.60 & 31.20& 69.60& 61.25& 42.00&44.75 &71.33 &38.95 &48.00 & 57.38\\
ST CoT (2 iters)&86.60 &31.40 &\textbf{72.20} &68.75  &48.00  &69.50  & 62.00 &54.74 &66.00 &64.23\\
ST CoT (3 iters)&87.00 &31.80 & 69.80 &68.00  &48.00  &69.25  & 58.00 &55.79 & 68.00&63.68\\
ST CoT (4 iters)&83.00 &30.00 & 66.40 &66.50  &43.00  &67.75  & 59.33 &50.53 & 66.00&61.20\\
ST CoT (5 iters)&83.20&	29.80&	63.00&	66.00&	45.00&	65.50&	58.00&	48.42&	66.00& 60.11\\
RM-Regen        &\textbf{87.20}&\textbf{34.80}&{71.40}&\textbf{73.00}&\textbf{52.00}&\textbf{77.75}&\textbf{79.33}&\textbf{56.84}&\textbf{73.00}&\textbf{68.05}\\
\bottomrule
\end{tabularx}
\caption{Accuracy on nine benchmarks without oracle verification. This table extends by additionally reporting results of 4 and 5 iterations. }
\label{tab:full without oracle verification}
\end{table*}

Table~\ref{tab:full without oracle verification} provides the full per-dataset
results, complementing Table~\ref{tab:without oracle verification}
in the main text. In particular, we report each iterative method with 2–4 iterations for both GPT-3.5 and 2-5 iterations LLaMA~3.1-8B, together with our methods.

We can observe that when verification signals are noisy, increasing the number of verification-and-rectification iterations does not guarantee consistent performance gains. Across multiple baselines, additional iterations often lead to diminishing returns and, in some cases, performance degradation, indicating that accumulated verification errors can adversely affect downstream rectification.
Our method consistently achieves the strongest or near-strongest performance across most benchmarks and both base models under noisy verification. This suggests that an RM-guided verification-and-regeneration strategy is more robust to verification noise and more efficient for performance improvement than repeatedly applying verification-and-rectification loops.

\section{Further Analysis}
\label{sec:further analysis}

\subsection{Ablation Study}
\label{sec:ablation}

\noindent
\textbf{Inference configuration}~ Table~\ref{tab:ablation-inference} compares different inference configurations. Our full setting, combining retrieval, verification, and regeneration, achieves the best accuracy of 64.00\%. Using fixed demonstrations without retrieval leads a substantial drop 9.4\%, highlighting the importance of query-dependent retrieval. Direct performance boost without verification and regeneration performs the worst, suggesting that simply adding RM context is insufficient. Rectification instead of regeneration lags behind by 10.40\%. 
These results confirm that retrieval, verification, and regeneration play complementary roles, and that our full configuration offers the best performance.

\begin{table}[htbp]
\centering
\scriptsize
\begin{tabularx}{\columnwidth}{l *{4}{c}}
\toprule
Variant & Retrieval & Verification & Regeneration  & Acc. (\%) \\
\midrule
Full (Ours)           & \cmark & \cmark & \cmark & 64.00\\
Fixed demos           & \xmark & \cmark & \cmark & 54.60\\
Direct boost          & \cmark & \xmark & \xmark & 44.60\\
Rectification         & \cmark & \cmark & \xmark & 53.60\\
\bottomrule
\end{tabularx}
\caption{GPT-3.5 on GSM\_Hard: ablation of inference-stage configurations.}
\label{tab:ablation-inference}
\end{table}

\noindent\textbf{RM curation}~ Table~\ref{tab:ablation-curation} shows that all three curation steps contribute to the final performance under oracle verification. Removing the teacher model causes the largest drop, from 68.60\% to 56.80\%, indicating that high-quality teacher feedback is crucial for constructing effective reflection memories. Disabling the ICL-oriented refine step leads to a moderate decrease of 5.4\%, suggesting that adapting raw reflections and rectifications into an ICL-friendly format is important. Without this refinement, even correct rectifications may be suboptimal as in-context demonstrations.
Furthermore, skipping the filter is less harmful but still hurts performance 2.2\%, implying that filtering primarily serves to remove low-quality or redundant memories and thus improves robustness rather than driving the main gains. 
Overall, these results suggest that denoising, filtering, and teacher-guided correction are all important, with teacher guidance being the dominant factor.

\begin{table}[htbp]
\centering
\scriptsize
\begin{tabularx}{0.87\columnwidth}{l *{4}{c}}
\toprule
Variant & Refine (ICL) & Filter & Teacher & Acc. (\%) \\
\midrule
Full (Ours)    & \cmark & \cmark & \cmark &  68.60\\
w/o Refine    & \xmark & \cmark & \cmark  &  63.20\\
w/o Filter     & \cmark & \xmark & \cmark &  66.40\\
w/o Teacher    & \cmark & \cmark & \xmark &  56.80\\
\bottomrule
\end{tabularx}
\caption{GPT-3.5 on GSM\_Hard: ablation of RM curation components (Refine, Filter, Teacher) under oracle verification.}
\label{tab:ablation-curation}
\end{table}

\noindent
\textbf{RM contents}~ Table~\ref{tab:ablation-components} examines which components of the RM are most beneficial. Removing either validate solution demonstrations $\mathcal{R}^{+}$ or reflections-and-rectification demonstrations $\mathcal{R}^{-}$
yields 4.2-4.4\% drops, showing that both corrected and contrastive erroneous traces are necessary for strong in-context guidance. In contrast, removing distilled principles slightly degrades performance, indicating that principles provide an additional but secondary gain compared to explicit positive/negative examples.
Overall, the ablation highlights the complementary nature of RM components: instance-level positive and negative examples form the core signal, while distilled principles provide higher-level generalization cues that further improve reasoning behavior.

\begin{table}[htbp]
\centering
\scriptsize
\begin{tabularx}{0.81\columnwidth}{l *{4}{c}}
\toprule
Variant & $\mathcal{R}^{+}$ & $\mathcal{R}^{-}$ & Principle  & Acc. (\%) \\
\midrule
Full (Ours)           & \cmark & \cmark & \cmark & 68.60\\
w/o $\mathcal{R}^{+}$ & \xmark & \cmark & \cmark & 64.40\\
w/o $\mathcal{R}^{-}$  & \cmark & \xmark & \xmark& 64.20\\
w/o Principle         & \cmark & \cmark & \xmark & 67.80\\
\bottomrule
\end{tabularx}
\caption{GPT-3.5 on GSM\_Hard: ablation of RM contents ($\mathcal{R}^+$, $\mathcal{R}^-$, principles) under oracle verification.}
\label{tab:ablation-components}
\end{table}

\subsection{Effect of RM Size on Performance}
\begin{figure}[htbp]
    \centering
    \includegraphics[width=0.9\columnwidth]{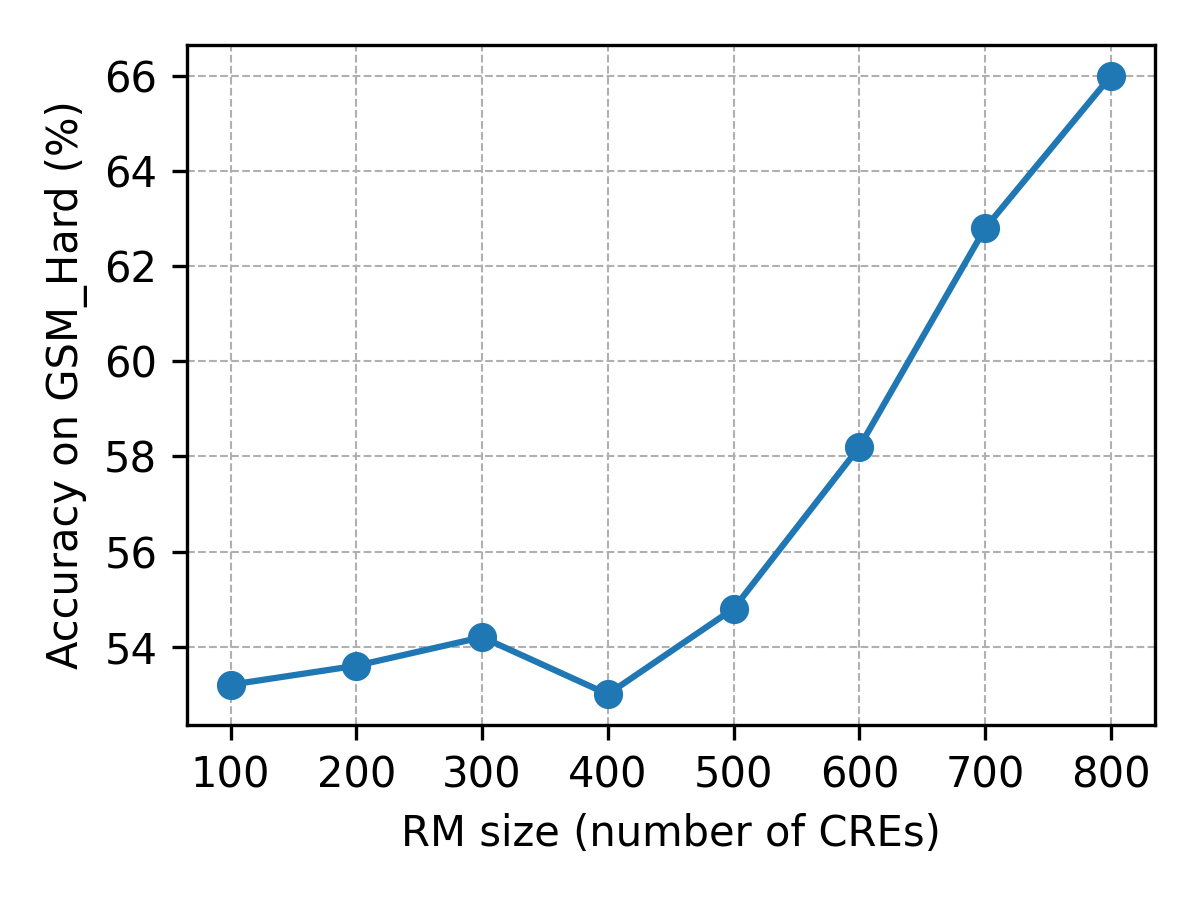}
    \caption{Effect of reflection-memory size on GPT-3.5 performance on GSM\_Hard.}
    \label{fig:rm-size-gsmhard}
\end{figure}
\label{sec:RM size}
Figure~\ref{fig:rm-size-gsmhard} shows the impact of reflection-memory size on GPT-3.5 accuracy on GSM\_Hard. When the RM is small (100–400 entries), performance fluctuates around 53–54\%, which is expected since limited memories provide sparse coverage and retrieval becomes sensitive to sampling noise and occasional mismatches. Accuracy increases more noticeably once the RM size reaches 500 (54.8\%) and continues to grow with larger memories, reaching 62.8\% at 700 and 66.0\% at 800 entries. The largely monotonic improvement in the large-memory regime suggests a coverage threshold beyond which additional contrastive reflections more reliably match the model’s error modes and supply useful rectification strategies. While larger RM may increase retrieval and context-construction cost, these results indicate that richer contrastive memories are crucial for challenging algorithmic reasoning on GSM\_Hard. Due to the limited dataset size, we did not evaluate larger RM sizes, and expect the gains to eventually saturate as RM grows further.

\subsection{Regeneration Rate and Flip Analysis}
\label{sec:wrong-to-correct}
Regeneration is triggered when the verifier determines the output to be incorrect. 
Therefore, the regeneration rate is equivalent to the rate of predictions verified as incorrect.

\textbf{With Oracle Verifier}~
When the verifier is an oracle, we analyze the rate of wrong-to-correct flips to better understand the effectiveness of our method. The evaluation of regeneration flip under an oracle verifier is shown in Table~\ref{tab:Regeneration correctness transition (oracle)}. On the challenging GSM-Hard dataset, our method successfully rectifies 46.55\% of the initial errors with the guidance of RM, substantially higher than the 10.47\% achieved by Reflexion. This result highlights the importance of incorporating external guidance context, which helps the model correct errors that it could not reliably resolve through iterative self-refinement alone.
\begin{table}[htbp]
\centering
\scriptsize
\begin{tabularx}{0.9\columnwidth}{l *{5}{c}}
\toprule
 &  &\multicolumn{2}{c}{RM-Regen} &\multicolumn{2}{c}{Reflexion (5 iters)}\\
 &  & Correct & Wrong & Correct & Wrong  \\
\midrule
\multirow{2}{*}{\rotatebox{90}{Initial}}
&Correct &{100.00\%}  & 0.00\% &{100.00\%}  & 0.00\% \\
&Wrong &\cellcolor{highGreen}\textbf{46.55\%}  & \cellcolor{lowRed}{53.45}\% &\cellcolor{lowGreen}{10.47\%}  & \cellcolor{highRed}89.53\%\\
\bottomrule
\end{tabularx}
\caption{GPT-3.5 on GSM-Hard: Regeneration correctness flip (with oracle verification)}
\label{tab:Regeneration correctness transition (oracle)}
\end{table}

\begin{table}[htbp]
\centering
\scriptsize
\begin{tabularx}{0.9\columnwidth}{l *{5}{c}}
\toprule
 &  &\multicolumn{2}{c}{RM-Regen} &\multicolumn{2}{c}{ST CoT (4 iters)}\\
 &  & Correct & Wrong & Correct & Wrong  \\
\midrule
\multirow{2}{*}{\rotatebox{90}{Initial}}
&Correct &\cellcolor{highGreen}\textbf{95.52\%}  & \cellcolor{lowRed}4.48\% &\cellcolor{midGreen}\textbf{86.05\%}  & \cellcolor{midRed}13.95\% \\
&Wrong &\cellcolor{highGreen}\textbf{38.63\%}  & \cellcolor{lowRed}61.37\% &\cellcolor{lowGreen}\textbf{8.77\%}  & \cellcolor{midRed}91.23\%\\
\bottomrule
\end{tabularx}
\caption{GPT-3.5 on GSM-Hard: Regeneration correctness flip (without oracle verification)}
\label{tab:Regeneration correctness transition (no oracle)}
\end{table}

\textbf{Without Oracle Verifier}~
When the verifier is not an oracle, we analyze both wrong-to-correct and correct-to-wrong flips to evaluate the effectiveness and robustness of our method. The evaluation result is shown in Table~\ref{tab:Regeneration correctness transition (no oracle)}. Our method maintains a strong error-correction capability, achieving a 38.63\% wrong-to-correct flip rate, 29.86\% higher than ST CoT (4 iters). Regarding robustness, we observe a marginal 4.48\% correct-to-wrong degradation, whereas ST CoT (4 iters) has 13.95\% degradation. Overall, the improvement and robustness confirm that our framework provides a reliable reasoning safety net, even when the internal verifier is imperfect.

\subsection{Generalizability of RM Across Models and Datasets}
\label{sec:out-of-domain}
\begin{table}[htbp]
\centering
\scriptsize
\begin{tabularx}{0.92\columnwidth}{l *{3}{c}}
\toprule
             & \textbf{GSM8K} & \textbf{GSM-Hard} & \textbf{MATH}  \\
\midrule
\textbf{GPT-3.5}      &90.00 \textcolor{gray!60}{/90.20} & 51.20 \textcolor{gray!60}{/68.60}& 66.00 \textcolor{gray!60}{/63.40}\\
\textbf{Llama 3.1-8B} &91.60 \textcolor{gray!60}{/92.60} & 35.00 \textcolor{gray!60}{/41.00}& 80.60 \textcolor{gray!60}{/80.60} \\
\textbf{Gemma 2-9B}   &91.40 \textcolor{gray!60}{/92.00} & 52.20 \textcolor{gray!60}{/53.00}& 71.20 \textcolor{gray!60}{/72.20}\\
\bottomrule
\end{tabularx}
\caption{Transferability of RM constructed for LLaMA-3.1-8B on MATH across models and datasets. Gray numbers denote native performance (the same model and dataset used for RM construction), while black numbers indicate cross-scenario performance under model or dataset shifts.
}
\label{tab:transfer}
\end{table}

Since RM construction incurs additional computational overhead, it is important to evaluate whether a single RM generalizes beyond the model and dataset used during its construction. We therefore examine two forms of out-of-domain transfer: (i) cross-model transfer, where an RM built for a source model is reused by different target models, and (ii) cross-dataset generalization, where RM signals derived from one dataset are applied to unseen tasks.
As shown in Table~\ref{tab:transfer}, applying RM across different models or tasks leads to a modest performance drop, since the distilled reasoning principles originate from the source model’s traces. Nevertheless, the transferred RM maintains strong performance across models and datasets, suggesting that it captures reusable reasoning patterns rather than overfitting to a specific model or task.
These results indicate that a single pre-constructed RM can support multiple models and tasks, making the framework practical for large-scale deployment where rebuilding RM for every model or dataset would be costly.

\section{Prompts}
\label{sec:prompts}

This section presents the prompt templates used in our method, including prompts for reasoning-quality scoring, response reflection, response verification, and principle extraction. 
These templates standardize the evaluation and refinement steps, ensuring consistent judgments and enabling reliable construction and retrieval of reflection memories across tasks.

\subsection{Principle Prompt}
\label{sec:principle prompt}
{\small
\begin{tcolorbox}[
    colback=gray!5,
    colframe=black!60,
    boxrule=0.5pt,
    arc=2pt,
    width=0.95\linewidth,
    left=5pt,
    right=5pt,
    top=4pt,
    bottom=4pt
]
\textbf{Developer:} You are an Insight Extraction Agent. \\

\textbf{Given:} \\
Reflection: \{analysis\} (contains detailed error analysis and corrective reasoning)\\

\textbf{Instructions:}
\begin{itemize}
\setlength{\itemsep}{1pt}      % 每个 \item 之间的垂直间距
\setlength{\parskip}{1pt}      % 段落之间
  \item Focus on deriving a concise, generalizable principle that can improve future reasoning performance.
  \item Do not restate or re-analyze the reflection; instead, abstract its key lessons into clear, reusable insights.
  \item Make the principle clear and actionable.
  \item The principle should be under 20 words.
\end{itemize}
\vspace{2pt}

\textbf{Output Format:}\\
\textless one concise principle or guideline that should be emphasized to improve future performance\textgreater
\end{tcolorbox}
}

\subsection{Reflection Prompt}
\label{sec:reflection prompt}
{\small
\begin{tcolorbox}[
    colback=gray!5,
    colframe=black!60,
    boxrule=0.5pt,
    arc=2pt,
    width=0.95\linewidth,
    left=5pt,
    right=5pt,
    top=4pt,
    bottom=4pt
]
\textbf{Developer:} You are a precise tutor. Analyze the errors in the original reasoning, then provide a concise and corrected reasoning.\\

\textbf{Given:}\\
{Question:} \{question\}\\
{Incorrect Reasoning:} \{prediction\}\\
{Gold Reasoning for Reference:} \{oracle\}\\[2pt]

\textbf{Instructions:}
\begin{itemize}
  \item In \texttt{Analysis}: Briefly list the main error(s) in the original reasoning  (1--3 bullet points max).
  \item In \texttt{Corrected Reasoning}:
    \begin{enumerate}
      \item Do \textbf{not} copy the original reasoning; fix it.
      \item Number the steps; include key equations and unit/constraint checks.
      \item Keep reasoning concise (no redundant explanations).
      \item Keep the final answer exactly equal to Gold Reasoning.
      \item Do not copy text verbatim from the Original Reasoning; rewrite with corrections.
    \end{enumerate}
\end{itemize}
\vspace{2pt}

\textbf{Output Format:}\\
Analysis: 
- \textless key mistake(s) \textgreater\\
Corrected Reasoning: 
\textless revised reasoning \textgreater
\#\#\#\# \textless final answer \textgreater
\end{tcolorbox}
}

\subsection{Verification Prompt}
\label{sec:verification prompt}
{\small
\begin{tcolorbox}[
    colback=gray!5,
    colframe=black!60,
    boxrule=0.5pt,
    arc=2pt,
    width=0.95\linewidth,
    left=5pt,
    right=5pt,
    top=4pt,
    bottom=4pt
]
\textbf{Developer:} You are an answer verification agent. Your task is to determine whether the provided answer is numerically and logically correct with respect to the question. \\

\textbf{Instructions:}
\begin{itemize}
\setlength{\itemsep}{1pt}      % 每个 \item 之间的垂直间距
\setlength{\parskip}{1pt}      % 段落之间
\item Extract the final answer.
\item Briefly check whether the answer satisfies the numeric or logical constraints of the question.
\item Focus on correctness, not writing style or explanation quality.
\end{itemize}
\vspace{2pt}

\textbf{Here are some assessment examples:} \\
Each example includes a question, an answer, an analysis of the answer, and a final assessment ("correct" or "incorrect").
Learn from these examples how to assess the correctness of reasoning.\\
\{examples\}\\

\textbf{Now assess the following case (do not repeat the answer in the analysis):} \\
Question: \{question\}\\
Answer: \{answer\}\\

\textbf{Output Format:}\\
Analysis: \textless brief reason \textgreater\\
Assessment: \textless correct/ incorrect \textgreater\\
\end{tcolorbox}
}

\subsection{ICL Oriented Refine Prompt}

\label{sec:refine prompt}
{\small
\begin{tcolorbox}[
    colback=gray!5,
    colframe=black!60,
    boxrule=0.5pt,
    arc=2pt,
    width=0.95\linewidth,
    left=5pt,
    right=5pt,
    top=4pt,
    bottom=4pt
]
\textbf{Developer:} You are a careful wrong example cleaner for contrastive In-Context Learning.\\
Contrastive In-Context Learning works by contrasting the wrong reasoning with the corrected reasoning.
Your task is to rewrite a noisy or verbose wrong example into a SHORT, CLEAR negative example that highlights exactly the mistake described in the Analysis.\\

\textbf{Given:}\\
{Question:} \{question\}\\
{Incorrect Reasoning:} \{prediction\}\\
{Analysis of the Mistake:} \{analysis\}\\
{Corrected Reasoning:} \{corrected\}\\[3pt]

\textbf{Instructions:}
\begin{itemize}
\setlength{\itemsep}{1pt}      % 每个 \item 之间的垂直间距
\setlength{\parskip}{1pt}      % 段落之间
\item Keep mistakes indicated in Analysis and preserve the WRONG final answer if present.
\item  Be as concise as the corrected reasoning.
\item No new variables, new inequalities, no new algebraic chains, no filler text, no paraphrasing.
\item Do NOT include any explanation or meta text—output ONLY the cleaned negative example.
\item If there is a wrong result, end the output with '\#\#\#\# \textless final answer \textgreater'.
\end{itemize}
\vspace{3pt}

\textbf{Output Format:}\\
Return ONLY the cleaned negative example with the wrong conclusion (if the original had one). End with the wrong answer in the format '\#\#\#\# \textless final answer \textgreater' if applicable.
\end{tcolorbox}
}

\subsection{Score Reasoning Quality Prompt}

\label{sec:score prompt}
{\small
\begin{tcolorbox}[
    colback=gray!5,
    colframe=black!60,
    boxrule=0.5pt,
    arc=2pt,
    width=0.95\linewidth,
    left=5pt,
    right=5pt,
    top=4pt,
    bottom=4pt
]
\textbf{Developer:} You are a strict grader for reasoning quality.
Evaluate if the reasoning is high-quality for use as an ICL demonstration.\\

\textbf{Given:}\\
{Question:} \{question\}\\
{Reasoning:} \{prediction\}\\[2pt]

\textbf{Instructions:}\\
Score the original reasoning strictly in [1–10] based on:
\begin{itemize}
\setlength{\itemsep}{1pt}      % 每个 \item 之间的垂直间距
\setlength{\parskip}{1pt}      % 段落之间
\item Logic validity: reasoning is logically sound, no gaps.  
\item Arithmetic \& units: calculations and units are correct.  
\item Completeness \& checkability: key steps are covered and can be verified.  
\item Brevity \& clarity: concise and clear without redundancy, similar to the gold reasoning.  
\item Format compliance: proper step-by-step to final answer structure. 
\end{itemize}
\vspace{2pt}

\textbf{Scoring anchors:}\\
\begin{itemize}
\setlength{\itemsep}{1pt}      % 每个 \item 之间的垂直间距
\setlength{\parskip}{1pt}      % 段落之间
\item 1–3 = unusable (major errors, misleading)  
\item 4–6 = partially usable (some errors but partial reasoning correct)  
\item 7–8 = good (minor flaws but mostly correct and clear)  
\item 9–10 = excellent (fully correct, concise, clear, ICL-ready)  
\end{itemize}
\vspace{2pt}

{Output strictly a single integer (1–10). Do not add any words, symbols, or explanations.}\\
\end{tcolorbox}
}

\end{document}